\documentclass[./sn-basic,Numbered]{./sn-jnl}
\usepackage{comment} 
\usepackage{subcaption}
\usepackage{siunitx}
\usepackage{booktabs}
\usepackage{etoolbox}
\usepackage{mathtools}

\DeclarePairedDelimiter\floor{\lfloor}{\rfloor}
\usepackage{graphicx}      
\newcommand{\Tau}{\mathrm{T}}
\usepackage{makecell}
\usepackage{hyperref}
\usepackage{cleveref}[2012/02/15]

\usepackage{algorithm}
\usepackage{amsmath,amsfonts}
\usepackage{algorithm}
\usepackage{array}
\usepackage{textcomp}
\usepackage{stfloats}
\usepackage{url}
\usepackage{verbatim}
\usepackage{graphicx}
\usepackage{cite}

\usepackage{graphicx}%
\usepackage{multirow}%
\usepackage{amsmath,amssymb,amsfonts}%
\usepackage{amsthm}%
\usepackage{mathrsfs}%
\usepackage[title]{appendix}%
\usepackage{xcolor}%
\usepackage{textcomp}%
\usepackage{manyfoot}%
\usepackage{booktabs}%
\usepackage{algorithm}%
\usepackage{algorithmicx}%
\usepackage{algpseudocode}%
\usepackage{listings}%

\theoremstyle{thmstyleone}%
\theoremstyle{thmstyletwo}%
\theoremstyle{thmstylethree}%
\begin{document}

\title[Memory-free Online Change Point Detection:\\ A Novel Neural Network Approach]{Memory-free Online Change Point Detection:\\ A Novel Neural Network Approach}

\author*[1]{\fnm{Zahra} \sur{Atashgahi}}\email{z.atashgahi@utwente.nl}
\author[3, 2, 1]{\fnm{Decebal Constantin} \sur{Mocanu}}\email{decebal.mocanu@uni.lu}
\author[1]{\fnm{Raymond} \sur{Veldhuis}}\email{r.n.j.veldhuis@utwente.nl}
\author[2]{\fnm{Mykola} \sur{Pechenizkiy}}\email{m.pechenizkiy@tue.nl}

\affil*[1]{\orgdiv{Faculty of Electrical Engineering, Mathematics and Computer Science}, \orgname{University of Twente}, \orgaddress{\country{the Netherlands}}}

\affil[2]{\orgdiv{Department of Mathematics and Computer Science}, \orgname{Eindhoven University of Technology}, \orgaddress{ \country{the Netherlands}}}

\affil[3]{\orgdiv{Department of Computer Science}, \orgname{University of Luxembourg}, \orgaddress{\country{Luxembourg}}}

\abstract{Change Point Detection (CPD), which detects abrupt changes in the data distribution, is recognized as one of the most significant tasks in time series analysis. Despite the extensive literature on offline CPD, unsupervised online CPD still suffers from major challenges, including scalability, hyperparameter tuning, and learning constraints. To mitigate some of these challenges, in this paper, we propose a novel deep learning approach for unsupervised online CPD from a multi-dimensional time series, named \textbf{A}daptive \textbf{L}STM-\textbf{A}utoencoder \textbf{C}hange  \textbf{P}oint \textbf{D}etection (\textbf{ALACPD}). ALACPD exploits an LSTM-autoencoder-based neural network to perform unsupervised online CPD. It continuously adapts to the incoming samples without keeping the previously received input, thus being memory-free. We perform an extensive evaluation on several real-world time series CPD benchmarks. We show that ALACPD, on average, ranks first among state-of-the-art CPD algorithms in terms of quality of the time series segmentation, and it is on par with the best performer in terms of the accuracy of the estimated change points. The implementation of ALACPD is available online on Github\footnote{\url{https://github.com/zahraatashgahi/ALACPD}}.}


\maketitle

\section{Introduction}
Time series analysis is increasingly set to become a vital task in many fields such as medicine, finance, and industry \cite{hamilton2020time}. change point detection (CPD) refers to the problem of finding abrupt changes in the behavior of the system \cite{basseville1993detection, polunchenko2012State, aminikhanghahi2017survey, lu2019fuzzy}. It is among the most significant tasks in time series analysis since change points contain vital information about the underlying data generation process. CPD has a wide range of application domains such as healthcare \cite{gee2018bayesian, muggeo2011efficient, malladi2013online, theocharopoulos2023analysing}, human activity monitoring \cite{villarroel2017non}, industrial systems \cite{shi2021dual}, financial data analysis \cite{lavielle2007adaptive}, and climate modeling \cite{reeves2007review}. It has been also recently used to detect changes in tasks in continual learning framework \cite{rao2019continual, li2021detecting}.

In this paper, we focus on unsupervised online CPD from multivariate time series, which is a crucial task in many real-world applications \cite{aminikhanghahi2017survey}. Online CPD algorithms process data as soon as they are available, unlike offline methods that detect changes after the entire dataset is collected \cite{truong2020selective}. Online CPD methods aim at finding change points with minimum delay after they happen. In addition, since providing annotations is a laborious task in many real-world problems, performing unsupervised CPD is of great interest in such applications. Finally, with the emergence of big data in recent years, there are many applications where data is constantly collected from multiple sources. Therefore, it is of great importance to design CPD algorithms that can process multi-dimensional time series \cite{wei2018multivariate}.

Despite the importance of unsupervised online CPD, current solutions still suffer from various challenges. A key problem with much of the literature is being highly dependent upon the choice of hyperparameters. Most existing CPD algorithms require hyperparameter-tuning to achieve their highest potential performance, e.g., Bayesian online change point detection (BOCPD) \cite{adams2007bayesian, van2020evaluation}. However, performing hyperparameter-tuning might not be feasible in many real-world online learning applications. Continuous data streams with evolving characteristics may require frequent adjustments to hyperparameters, disrupting the online learning process. Additionally, obtaining labeled data for tuning is not always practical or cost-effective in unsupervised CPD scenarios. Time constraints and the need for rapid decision-making further hinder the lengthy process of hyperparameter optimization in dynamic online learning scenarios.

Another major drawback regarding a vast majority of previous works is requiring prior information about data \cite{aminikhanghahi2017survey}. This prior information can be regarding various aspects, including data distribution (e.g., stationary, i.i.d), change points (e.g., number of change points), and states' characteristics (state is a portion of time series between two consecutive change points). Therefore, these algorithms are impractical in online real-world scenarios where no information about the system is available prior to the change point detection phase. Finally, some existing online CPD methods require saving the entire or a part of the input data stream. This can lead to high memory requirements, which might be infeasible in low-resource devices on edge. These challenges lead us to conclude that unsupervised online CPD remains a significant open challenge in real-world scenarios.

\begin{figure}[!t]
        \begin{center}
        \centerline{\includegraphics[width=\columnwidth]{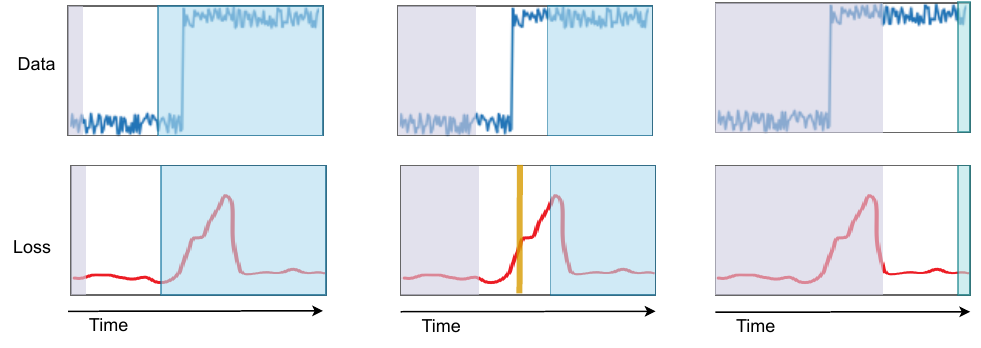}}
        \caption{Overview of ALACPD. gray shows the previously received data, white is the current data window and blue is the future unseen data. ALACPD exploits the reconstruction loss of the data (red line) to detect change points. If the reconstruction loss for the samples goes beyond the average loss for several time steps in a row (middle), ALACPD reports a change point and adapts itself to the new data distribution (right).}
        \label{fig:schematic_approach}
        \end{center}
    \end{figure}   
Deep neural networks, particularly recurrent neural networks (RNNs), have shown great success in time series analysis, including classification \cite{ismail2019deep}, forecasting \cite{hewamalage2021recurrent}, and anomaly detection \cite{cook2019anomaly}. This is mainly due to their ability to capture complex non-linear representations from multivariate data. In addition, while statistical machine learning methods might fail in processing large-scale data, deep learning models' performance scales with the dataset size \cite{hewamalage2021recurrent}. Despite the advantages of deep neural networks compared to traditional machine learning methods, there are only very few works that have used these models to perform CPD \cite{chang2018kernel, deldari2021time, de2021change, du2021finder, ebrahimzadeh2019deep}. This is because CPD using deep learning techniques suffers from various challenges, such as heavy demands on large labeled datasets \cite{cini2022filling}, high memory requirements, and low performance in online time series processing.

To address some of these challenges, this paper presents a novel deep learning-based approach to the problem of unsupervised online CPD. Our proposed algorithm, \textbf{A}daptive \textbf{L}STM-\textbf{A}uto\textbf{e}ncoder \textbf{C}hange-\textbf{p}oint \textbf{D}etection (\textbf{ALACPD}), seeks to detect changes from multi-dimensional time series in an unsupervised manner using minimum memory resources. Specifically, our contributions are:
\begin{itemize}
    \item We design a novel autoencoder architecture, named \textbf{T}ime series \textbf{A}uto-\textbf{E}ncoder \textbf{net}work (TAEnet), for learning short-term and long-term dependencies from multi-dimensional time series. Our designed architecture consists of an LSTM-autoencoder network with skip connections and an auto-regressive block. This architecture is presented in Figure \ref{fig:TAEnet}.
    \item Using the designed architecture, we introduce a novel algorithm for the problem of online CPD, named ALACPD. Our proposed algorithm exploits the reconstruction loss of an ensemble of TAEnet networks to identify change points in the input time series. A simplified overview of our proposed approach is shown in Figure \ref{fig:schematic_approach}. ALACPD can detect multiple change points from a multivariate time series in an online unsupervised manner. As ALACPD exploits only the reconstruction loss of the data to predict changes, it does not impose any assumptions about the types of the detected changes. To our knowledge, this is the first work that exploits LSTM-Autoencoders to perform online unsupervised CPD.
    \item ALACPD is robust to anomalies; it can distinguish anomalies from change points. 
    \item ALACPD is memory-free; by discarding the input samples in a short time after receiving it, ALACPD has a low memory footprint.
\end{itemize}

\section{Background}
This section provides a brief overview of existing CPD algorithms and their categorization. Then, we describe recurrent neural networks and their applications in event detection from time series. 

\subsection{Change Point Detection}
Real-world time series data reveal non-stationary behavior where the mean, variance, and auto-correlation of the time series vary with time. Such behavior can be categorized as concept drift, seasonality, or change point. While concept drift and seasonality represent changes in the statistical distribution of time series that occur over time or cyclical, change points represent more abrupt changes than the earlier categories \cite{cook2019anomaly}.

There exists an extensive literature on CPD from time series. Existing CPD algorithms can be categorized into different types based on various criteria \cite{aminikhanghahi2017survey}: $(1)$ \textit{Processing delay}: CPD methods can be divided into online or offline methods. Offline methods \cite{truong2020selective} process the entire dataset at once, while online methods process data as soon as they enter the system. Online CPD methods have gained increasing popularity due to their ability to adapt to real-time data and the changing demands of various applications. This shift towards online CPD is primarily driven by the need for more functionality in real-world scenarios, such as real-time monitoring, dynamic decision-making, and continuous learning. In this paper, we propose an online CPD algorithm. $(2)$ \textit{Availability of labels}: Change points can be detected in a supervised, semi-supervised, or unsupervised manner. $(3)$ \textit{Data dimensionality}: based on the dimensionality of time series, CPD algorithms can be classified into univariate or multivariate algorithms. $(4)$ \textit{Algorithm assumption}: CPD methods can be categorized as parametric or non-parametric \cite{matteson2014nonparametric}, depending on the assumptions they make about the underlying data distribution \cite{van2020evaluation}.

Various techniques have been used to address the unsupervised CPD problem in the literature. One of the popular techniques to perform CPD is the Bayesian approach. While early Bayesian approaches perform CPD in an offline manner \cite{barry1993bayesian, truong2020selective, chib1998estimation}, they have been vastly investigated for online CPD \cite{fearnhead2007line, adams2007bayesian,maslov2017blpa, agudelo2020bayesian, gundersen2021active, alami2020restarted, li2021detecting}. Bayesian online change point detection (BOCPD) \cite{adams2007bayesian} is a popular Bayesian method to perform online CPD; it performs CPD by estimating the posterior probability over the run length (the time since the last change point). However, BOCPD makes strong assumptions about the underlying data distribution, and it is sensitive to the choice of hyperparameters \cite{van2020evaluation, ahmad2017unsupervised}. Moreover, the computational cost of this algorithm grows with the run-length \cite{Gregory2019bocpd}. Kernel-based methods have also been a widely used technique to perform CPD \cite{harchaoui2009kernel, NIPS2015_eb1e7832, arlot2019kernel, chang2018kernel}. They map samples to higher-dimensionality feature space and perform CPD by comparing the homogeneity of each subsequence. A drawback of most Kernel-based methods is their dependency on the choice of kernel function and its parameters \cite{chen2015graph}. Other techniques such as clustering \cite{keogh2001online}, subspace models \cite{kawahara2007change, moskvina2003algorithm}, Gaussian process \cite{saatcci2010gaussian}, CUSUM principle \cite{page1954continuous}, Graph-based models \cite{chen2015graph}, distance-based methods \cite{matteson2014nonparametric}, moving average \cite{frittoli2021change}, and density ratio models \cite{aminikhanghahi2018real} have been also used to detect change points from time series.


\subsection{Recurrent Neural Networks for Event Detection}
Recurrent neural networks (RNNs) are a type of neural network that has been designed to perform time series processing \cite{rumelhart1985learning}. By performing automatic feature engineering, deep neural networks generally learn a rich representation of data that is beneficial to the final task of interest. Long short-term memory (LSTM) networks are a strong variant of RNNs that can learn long-term data representations. LSTMs have shown successful application in diverse domains, including speech recognition \cite{graves2013hybrid}, sentiment analysis \cite{wang2016dimensional}, healthcare \cite{wang2018lstm, lipton2015learning}, and weather prediction \cite{karevan2020transductive}. LSTM networks excel at capturing long-term dependencies in data, enabling them to discern subtle patterns and anomalies, while their real-time adaptability further reinforces their suitability for dynamic decision-making and continuous monitoring. With the added advantage of reduced reliance on manual feature engineering, their robust feature extraction capabilities and superior performance demonstrate the potential for accurate and reliable change point detection.

One of the most important applications of RNNs for time series is anomaly detection. By learning long-term temporal representation of the sequence, LSTMs have shown great performance in anomaly detection \cite{lindemann2021survey, malhotra2016lstm, malhotra2015long, zhang2019deep, kieu2019outlier, MALEKI2021107443, liu2023new}. A popular approach for anomaly detection with LSTMs is to use the reconstruction loss of samples to detect anomalies. This category of models trains the network on anomaly-free data. During inference, they use a threshold mechanism on the reconstruction loss to detect anomalies. The samples with a reconstruction exceeding this threshold are reported as an anomaly. 

Only a few works have used recurrent neural networks to perform CPD. In \cite{ebrahimzadeh2019deep}, authors have proposed a novel RNN architecture, named pyramid recurrent neural network, to perform multi-scale supervised CPD. In \cite{du2021finder}, authors exploit a time series predictor module and a classifier to detect change points. However, they only focus on supervised change point detection. To the best of our knowledge, this research is the first work that proposes online unsupervised CPD using RNN-based networks. 

\section{Methodology}

In this section, we first start by describing the change point detection problem in Section \ref{ssec:problem_Setup}. Next, in Section \ref{ssec:TAEnet}, we present the architecture of our introduced network for learning the representation of a given time series, named \emph{TAEnet}. Finally, in Section \ref{ssec:proposed_method}, we describe how TAEnet is exploited to develop our proposed CPD algorithm, \emph{ALACPD}. 

\subsection{Problem Setup}\label{ssec:problem_Setup}
\paragraph{Notations} 
Throughout this paper, we use the following terms and notations. $T = \{x_1, x_2, ..., x_n\}$ denotes a time series, where $x_t \in \mathbb{R}^D$ is the observation at time $t$. If $D = 1$, $T$ is called a one-dimensional time series, and if $D > 1$, $T$ is a multi-dimensional time series. We denote a segment of time series between time steps $i$ and $j$ as $x_{i:j}$. By using an sliding window of length $\mathit{w}$ across $T$, $X_t = \{x_{t-w+1}, x_{t-w+2}, ..., x_t\}$ is a sub-sequence of $T$. We consider $X_t$ as the input of the model at time $t$. We call $X_{i:j} = \{X_{i}, X_{i+1}, ...,X_j\}$ as an interval of $T$ between time steps $i$ and $j$. 

\paragraph{Definitions} 
Change point is defined as a time step in time series $T$ where the properties of $T$ abruptly change. Time series $T$ contains $K$ change points $ \mathcal{T} = \{ \tau_i | i \in \{1,2, ..., K\}, \tau_i \in \{1,2, ..., n\}\}$. The set of change points $\mathcal{T}$ gives a partitioning over the time series $T$, $\mathcal{P}$, which segments $T$ into disjoint sets (called state) $\mathcal{A}_i = x_{\tau_{i-1}:\tau_{i}}$, where $i \in \{1,2, ..., K\}\}$. In summary, $\Tau$ segments time series into disjoint states, such that $\mathbb{P}_{X_i} = \mathbb{P}_{X_j}$ if $X_i$ and $X_j$ are in the same state, and $\mathbb{P}_{X_i} \neq \mathbb{P}_{X_j}$ if $X_i$ and $X_j$ are in two subsequent states. $\mathbb{P}_{X_i}$ is the probability density function of $X_i$. We also denote $\mathbb{P}_t$ as the probability density function of the current state.

\paragraph{Problem Definition} 
Change point detection (CPD) is the problem of approximating the set of change points $\mathcal{T} $. We assume that the number of change points $K$ is unknown.

\subsection{TAEnet Architecture}\label{ssec:TAEnet}
In this section, we present our proposed autoencoder-based network for learning time series, named \textbf{T}ime series \textbf{A}uto\textbf{E}ncoder \textbf{net}work (TAEnet), which will be used in Section \ref{ssec:proposed_method} to introduce our proposed CPD algorithm. 

\begin{figure*}[!t]
        \begin{center}
        \centerline{\includegraphics[width=\textwidth]{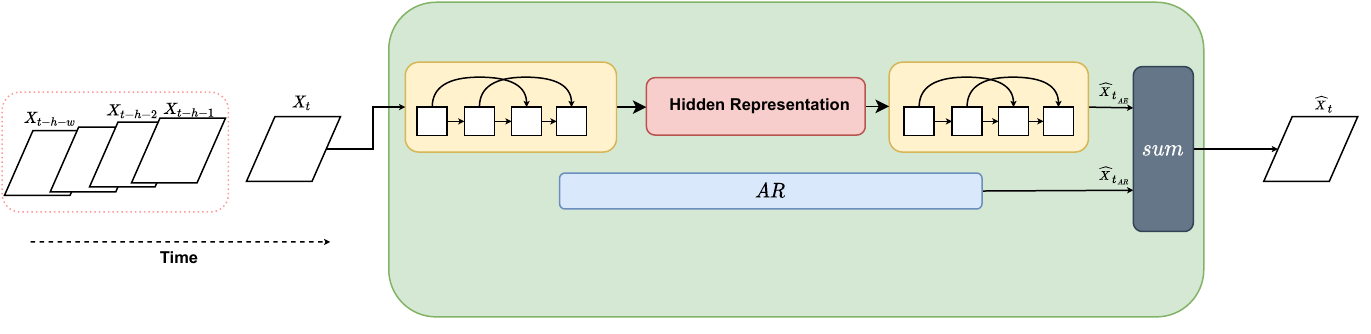}}
        \caption{TAEnet architecture.}
        \label{fig:TAEnet}
        \end{center}
    \end{figure*}

In short, TAEnet reconstructs the input time series using two main components, including an LSTM autoencoder-based network and an auto-regressive function. The architecture of this network is shown in Figure \ref{fig:TAEnet}. In the following, we first describe each of these two components in Sections \ref{sssec:recurrent_component} and \ref{sssec:AR_component}. Finally, in Section \ref{sssec:TAEnet}, we present TAEnet and its objective function.

%
%

\subsubsection{Recurrent Component}\label{sssec:recurrent_component}
Our proposed recurrent component which is an LSTM-Autoencoder-based network, is designed to learn the short- and long-term dependencies in the input time series. We introduce a novel LSTM architecture, named \textbf{A}daptive \textbf{S}kip-\textbf{C}onnected \textbf{LSTM} (ASC-LSTM), which is inspired by the Skip-Connected LSTM (SC-LSTM) \cite{wang2016recurrent}. In \cite{wang2016recurrent}, it has been shown that residual learning significantly improves LSTM’s performance. We use ASC-LSTM instead of standard LSTM in the LSTM-Autoencoder network, to boost the learning, and learn dependencies at different time scales. In the following, we elaborate on the structure of the recurrent component.

SC-LSTM employs skip connections among the non-consecutive units of LSTM with distance $S$, to improve the flow of information in the network. Therefore, the hidden state for each unit will be computed as:

\begin{equation}
    h_t = f_1(c_t)\circ o_t + \alpha h_{t-S},
\end{equation}

where $h_t$ is the hidden state at time step $t$, $f_1$ is the activation function, $c_t$ is the cell memory, $o_t$ is the output, $\alpha$ is a scalar value that can be fixed or trained as a model parameter and $S$ is the skip size. In \cite{wang2016recurrent}, authors deploy the skip connection only in time steps $1, 1+S, 1+2S, ..., \floor{\frac{U-S-1}{S}}S$, where $U$ is the number of hidden units of the LSTM network.

In this paper, we introduce more flexibility in the structure of SC-LSTM. We employ skip connections at all the time steps and let the network find the best value for $h_t$, using the information of the current time step ($t$) and the hidden state of a cell in the time step $t-S$. Formally, we have:
\begin{equation}
    h_t = \alpha f_1(c_t)\circ o_t + (1-\alpha) f_2(h_{t-S}),
\end{equation}

where $\alpha$ is a scalar value between $0$ and $1$, and $f_2$ is the activation function for the skip connection. We deploy this skip connection at all time steps $1, 2, ..., U-S$. The network learns the proper values for $\alpha$ and the weight of skip connection at each time step during training. We call this network \textbf{A}daptive \textbf{S}kip-\textbf{C}onnected \textbf{LSTM} (ASC-LSTM). 

Our proposed recurrent component is an Autoencoder network which has ASC-LSTM networks in its encoder and decoder part; we call this network \textbf{ASC-LSTM} \textbf{A}uto\textbf{E}ncoder (\textbf{ASC-LSTM-AE}). At each time step it reconstructs the current input $X_t$ as $ \widehat{X}_{t_{AE}}$. By reconstructing the input time series in the output and exploiting the skip connections in the structure, ASC-LSTM-AE learns a useful representation of the data including the short- and long-term dependencies. 

\subsubsection{Auto-regressive (AR) Component}\label{sssec:AR_component} 
In addition to the recurrent component, which learns long- and short-term non-linear dependencies, we exploit an AR model to increase the sensitivity of the model to local scale changes in the input. In \cite{lai2018modeling}, authors have shown that using an AR model in combination with a recurrent neural network can significantly enhance the accuracy of time series prediction. In this work, we propose to deploy an AR model in parallel to the recurrent component to improve the quality of the time series reconstruction. This AR model predicts the current input $X_t$ as $ \widehat{X}_{t_{AR}}$ using the previously received input. This component can be formulated as follows:

\begin{equation}
    \widehat{x}_{t_{AR}}^d = \sum_{i=0}^{w-1}W^{AR}_{i}x_{t-h-i}^d + b,
\end{equation}

where $\widehat{x}_{t_{AR}}^d$ denotes the prediction of the AR model at time step $t$,  $d\in\{1, ..., D\}$ is the index for dimension, $W\in \mathbb{R}^w$ and $b\in \mathbb{R}$ are the coefficients of the AR model which are shared for all the dimensions, and $h$ is a hyperparameter in time series prediction problems which denotes the horizon for prediction. We will explain in Section \ref{ssec:ablation} why we use this horizon for change point detection problem. By predicting $\widehat{x}_{t^\prime_{AR}}$ for $t^\prime\in\{t-w+1, ..., t-1, t\}$ we can predict/reconstruct the current window of input as $ \widehat{X}_{t_{AR}}$.

\subsubsection{TAEnet}\label{sssec:TAEnet} 
Here, we describe our proposed architecture for learning time series, named \textbf{T}ime series \textbf{A}uto\textbf{E}ncoder \textbf{net}work (TAEnet). TAEnet exploits the introduced recurrent (Section \ref{sssec:recurrent_component}) and auto-regressive (Section \ref{sssec:AR_component}) components to reconstruct the input. By summing the output of these components, it reconstructs the input $X_t$ observation in the output, and minimizes its reconstruction loss as follows:

\begin{equation}\label{eq:reconstructing_x_t}
\widehat{X}_t =  \widehat{X}_{t_{AR}} + \widehat{X}_{t_{AE}},
\end{equation}

\begin{equation}\label{eq:objective_function}
\min \sum_{t=1}^{n}  \left\Vert X_t- \widehat{X}_t   \right\Vert_2^2,
\end{equation}

where $\widehat{X}_t$ is the reconstruction of input at time step $t$. By minimizing the reconstruction loss over the time series, TAEnet learns the short- and long-term dependencies in the time series. It should be noted that Equation \ref{eq:reconstructing_x_t} is a weighted sum of the outputs of the two components, and the network learns the weights during the training phase. In the next subsection, we will explain how we use TAEnet to perform CPD.

\subsection{Adaptive LSTM-Autoencoder Change Point Detection (ALACPD)}\label{ssec:proposed_method}
This section presents our proposed algorithm for estimating change points in time series. In short, our proposed algorithm, \textbf{A}daptive \textbf{L}STM-Autoencoder \textbf{C}hange-\textbf{P}oint \textbf{D}etection (\textbf{ALACPD}), uses an ensemble of TAEnets (introduced in Section \ref{ssec:TAEnet}) to estimate the locations of change points. ALACPD reports that a change point has been detected when it observes that the input distribution is changing; it detects this change by monitoring the quality of the time series reconstruction by the TAEnet ensemble. In the following, we first describe the architecture of the network used for CPD in Section \ref{ssec:TAEnet_ensemble}. Then, in Section \ref{ssec:CPD_approach}, we present our proposed CPD approach, ALACPD. Finally, in Section \ref{ssec:training}, we explain ALACPD and its training process in more detail. 


\subsubsection{TAEnet Ensemble}\label{ssec:TAEnet_ensemble}
Our final designed network for change point detection is an ensemble of $M$ TAEnet with different structures. The difference among these networks is imposed by using different skip connection sizes. Therefore, each TAEnet learns a unique representation of the time series with a different time scale, which is determined by the size of the skip connection. Each of the TAEnets in the ensemble architecture is trained individually on the input data. 

\subsubsection{Change Point Detection}\label{ssec:CPD_approach} 
In short, our proposed CPD algorithm, ALACPD, continuously monitors the quality of the reconstructed input by each network in the TAEnets Ensemble to perform CPD. We suggest that the reconstruction loss of each sub-network can be used to detect change points. If the majority of the sub-networks cannot accurately reconstruct the input time series for several time steps in a row, it means that the distribution of the input is changing, and therefore, a change point is detected. 
 
ALACPD is inspired by anomaly detection in time series using LSTM-autoencoder-based networks \cite{malhotra2016lstm}. In \cite{malhotra2016lstm}, authors suggest that the reconstruction loss of an LSTM-autoencoder network that is trained on normal data (time series without anomalies) can be used to detect anomalies. In this network, if the reconstruction error of a test sample goes beyond a threshold, this sample is identified as an anomaly. In other words, it does not belong to the normal data distribution. 
 
In this paper, we propose to use the reconstruction loss of an autoencoder-based network (TAEnet) to detect change points in a given time series. If TAEnet is trained on the data $X_{i:j}$ with distribution $\mathbb{P}_{X_{i:j}}$, it gives a low reconstruction loss for any $X_t$ that $\mathbb{P}_{X_t} {\displaystyle =\,} \mathbb{P}_{X_{i:j}}$, and a high loss for any $X_t$ that $\mathbb{P}_{X_t} \neq \mathbb{P}_{X_{i:j}}$. As a result, if the data-generating process goes under a change and the data distribution changes, the model will have a high reconstruction loss for any sample generated from the new distribution. We use this information to detect change points; we suggest that the time step where the reconstruction loss of the model suddenly increases and continues for several time steps can be reported as a change point.
 
ALACPD exploits an ensemble of TAEnets with different skip connection sizes to detect change points. Each TAEnet is trained individually on the input data. However, ALACPD decides a new sample does not belong to the current data distribution only when a majority of sub-networks report a high reconstruction loss. The high reconstruction loss is reported when the reconstruction loss for a sample goes beyond the threshold of sub-network $m\in\{1, 2, ..., M\}$ ($th_{state}^m$). This threshold is determined by the average loss of samples in the current state for sub-network $m$. At each time step, if the received input belongs to the current data distribution, the model updates itself by training all the sub-networks on this sample. On the other hand, if the sample does not belong to this distribution, the TAEnet Ensemble will not be updated. A change point is detected when ALACPD identifies several samples in a row that do not belong to the current data distribution. 
 
After a change point is detected, ALACPD adapts itself to the new distribution of the input data. All sub-networks will be trained on the samples from the new distribution. In other words, we are observing a new state in the input time series.

\subsubsection{Training}\label{ssec:training}
In this section, we describe the ALACPD algorithm and its training process in detail. In short, ALACPD starts with initializing the architecture and the parameters. Then, it starts online training by processing each sample as soon as it is received. At each time step, it decides if a change point has been observed or not. We will elaborate on each of these steps in the following. The pseudo-code describing the ALACPD algorithm is summarized in Algorithm \ref{alg:ALACPD}.

\paragraph{Initialization} 
In ALACPD, first, the TAEnet Ensemble is initialized and trained offline for a few epochs to be prepared for the online training. We initialize $M$ TAEnets with different skip connection sizes $S$. Next, each TAEnet is trained individually for $e_{init}$ epochs with a small proportion of samples as $X_{1:n_{init}}$ ($n_{init}$ is usually less than $10\%$ of the samples) that contains no change points. Each of these TAEnets learns the dependencies at a different scale.

After the initial training, we calculate the threshold $th_{state}^m$ for each TAEnet.
\begin{equation}\label{eq:threshold_init}
    th_{state_{init}}^m = CL_{avg_{init}}^{m},
\end{equation}

\begin{equation}
    L_{avg_{init}}^{m} = \frac{1}{n_{init}}\sum_{t=1}^{n_{init}}  \left\Vert X_t- \widehat{X}_t^m   \right\Vert_2^2,
\end{equation}
where $\widehat{X}_t^m$ is the reconstruction of $X_t$ by sub-network $m$, and $C > 1$ is a hyperparameter of our algorithm, which determines the size of the threshold. If we set coefficient $C$ to a small value (e.g., $1$), many samples will be identified as not belonging to the current distribution, while setting $C$ to a very large value (e.g., $10$) results in assigning most of the samples to the current distribution. 

\paragraph{Training}
After the initialization phase is finished, the online training starts. As also explained in Section \ref{ssec:CPD_approach}, in this phase, ALACPD processes each sample as soon as they are collected and decides if a change point is observed or not. 

At each time step, ALACPD determines whether the new sample $X_t$ belongs to the current state of the system or not, by simply comparing the reconstruction loss of a sample $X_t$ with $th_{state}^m$ for all sub-networks: 

\begin{equation}\label{eq:anomaly_decision}
  \begin{cases}
    \mathbb{P}_{X_t} {\displaystyle =\,} \mathbb{P}_{t}, & \text{if$  \sum\limits_{\substack{m=1 \\  L_t^m > th_{state}^m}}^{m=M} 1 <  \lceil \beta M \rceil$}\\
     \mathbb{P}_{X_t}\neq \mathbb{P}_{t}, &  \text{otherwise}
     
  \end{cases}
\end{equation}

\begin{equation}
    L_t^m= \left\Vert X_t- \widehat{X}_t^m   \right\Vert_2^2,
\end{equation}

\begin{algorithm}[!t]
     \caption{ALACPD}\label{alg:ALACPD}

    \begin{algorithmic}
        \State \textbf{Input}: Time series $T$, ensemble size $M$, skip connection sizes $S$, $n_{init}$, training epochs $e_{init}$, $e_{train}$, and $e_{re-init}$.
        \State \textbf{Output}: Estimated set of change points $\mathcal{T}$.
        \vspace{2.5mm}
        \State \textbf{I. Initialization} (Section \ref{ssec:training} - Initialization)
            \State 1. Initialize $M$ TAEnets with different skip connection sizes $S$. 
            \State 2. Train each TAEnet individually for $e_{init}$ epochs with few samples $X_{1:n_{init}}$. 
            \State 3. Calculate the threshold $th_{state}^m$ for each TAEnet (Equation \ref{eq:threshold_init}).
        \vspace{2.5mm}
        \State \textbf{II. Online Training} (Section \ref{ssec:training} - Training)
            \For{ each time step \texttt{$t \in \{n_{init},\dots, n\}$}}
                \If{ $\mathbb{P}_{X_t} {\displaystyle =\,} \mathbb{P}_{t}$}
                    \State 1. Train each TAEnet on sample $X_t$ for $e_{train}$ epochs.
                    \State 2. Update the thresholds using Equation \ref{eq:update_tresholds}.
                    \State 3. Discard the set of anomalous samples.
                \Else   
                    \State 1. Keep $X_t$ as an anomalous sample.
                    \State 2. Check if a change point is detected or not:
                        \If{ $ \mathbb{P}_{X_{i}} \neq \mathbb{P}_{t}, \;\;$ $\forall i\, \in\, \{ {t-n_{CPD},\,  ...,\, t-1, \,t}\}$}
                            \State 1. Report $t-n_{CPD}$ as a change point and add it to $\mathcal{T}$.
                            \State 2. Train each TAEnet individually for $e_{re-init}$ epochs on $X_{t-n_{cpd}:t}$.
                            \State 3. Re-initialize the thresholds using Equation \ref{eq:threshold_init} and $X_{t-n_{cpd}:t}$.
                        \EndIf 
                \EndIf 
            \EndFor
        \State Return $\mathcal{T}$.
    \end{algorithmic}
    
\end{algorithm}

where $\beta$ (with a value between $0$ and $1$) is the hyperparameter of the algorithm controlling the threshold, and $L_t^m$ is the reconstruction loss of $X_t$ for sub-network $m$. 

Then, the training will proceed based on the distribution of $X_t$:
\begin{itemize}
    \item $\mathbb{P}_{X_t} {\displaystyle =\,} \mathbb{P}_{t}$\\
    If $X_t$ belongs to the current distribution, ALACPD learns this sample by updating the TAEnet Ensemble and the thresholds $th_{state}^m$. The TAEnet Ensemble is updated by training each sub-network on sample $X_t$ for $e_{train}$ epochs. Then, the thresholds will by updated as follows:
    \begin{equation}\label{eq:update_tresholds}
        th_{state_{new}}^m = CL_{avg_{new}}^{m},
    \end{equation}
    \begin{equation}
        L_{avg_{new}}^m = \frac{n_{\mathbb{P}_{t}}L_{avg_{old}}^m + L_t^m }{n_{\mathbb{P}_{t}} + 1},
    \end{equation}
    where $n_{\mathbb{P}_{t}}$ is the number of samples belonging to the current state of the data.

    \item $\mathbb{P}_{X_t} \neq \mathbb{P}_{t}$\\
    If $X_t$ does not belong to the current data distribution, ALACPD examines whether a change point has occurred. ALACPD reports a change point only when it receives $n_{CPD}$ samples in a row from a different distribution:
    \begin{equation}
        t- n_{CPD} \in \mathcal{T}\;\text{if   }\;\mathbb{P}_{X_{i}} \neq \mathbb{P}_{t}, \;\; \text{\scalebox{0.85}{  	$\forall i\, \in\, \{ {t-n_{CPD}+1,\,  ...,\, t-1, \,t}\}$}}.
    \end{equation}
    Otherwise, $X_t$ might be an anomalous or out-of-distribution sample, not a change point.
    
    If a change point is detected, TAEnet Ensemble should forget the previously learned distribution and be trained on the new data $X_{t-n_{cpd}:t}$. Therefore, it should be reinitialized or trained long enough to completely forget the learned representation. In this paper, we train the networks for a long period ($e_{re-init}$ epochs) to forget the previously learned data. After training TAEnet Ensemble, the thresholds $th_{state}^m$ should be re-initialized similar to Equation \ref{eq:threshold_init}. Then, the training continues as normal by processing the next input sample.
    
    However, a change point is not detected, and ALACPD does not learn this sample; it simply ignores it and processes the next sample. This sample might be anomalous or out-of-distribution data that happens for a short period. On the other hand, it can be a potential change point that can not be confidently identified at the current time step. ALACPD waits until this behavior continues for a time window of size $n_{CPD}$ and then decides whether a change point has been observed. Otherwise, if this behavior does not continue for $n_{CPD}$ time steps, ALACPD discards the anomalous samples as soon as it receives a normal sample. Therefore, the maximum delay for reporting a change point is $n_{CPD}$. 
    
\end{itemize}

\section{Experiments and Results}
In this section, we evaluate our proposed CPD algorithm and compare it with several state-of-the-art CPD methods. We first describe the experimental setting in Section \ref{ssec:experiments_settings}. Then, we present and discuss the obtained results in Section \ref{ssec:results}. 

\subsection{Settings}\label{ssec:experiments_settings}
In this section, we describe the settings of our experiments, including the datasets used for evaluation, hyperparameter values, implementation details, and evaluation metrics.

\subsubsection{Datasets}\label{ssec:datasets} In our experimental setting, we focus on CPD from real-world multi-dimensional time series. We evaluate our proposed algorithm and state-of-the-art CPD methods on four real-world multi-dimensional time series as follows:

\begin{itemize}
    \item \textbf{Apple \cite{van2020evaluation}.} The daily closing price and volume of Apple stock collected from Yahoo Finance. 
    \item \textbf{Occupancy \cite{mirugwe2020accurate}.} Room occupancy dataset consisting of sensors' measurement installed in a room. These sensors include temperature, humidity, light, and CO$_{2}$ sensors.
    \item \textbf{Run\_log \cite{van2020evaluation}.} Measurement of pace and total distance of a runner using a training program.
    \item \textbf{Bee\_waggle \cite{oh2008learning}.} The position of the honey bee moving between three states: left turn, right turn, and waggle.
\end{itemize}

These datasets, along with their annotated change point locations, are presented in Figure \ref{fig:datasets}. The ground truth for these time series is provided by five expert human annotators \cite{van2020evaluation}. The vertical grey lines in Figure \ref{fig:datasets} on the time series represent the change points detected by human annotators; each annotator is shown as a separate color on the change points detected by the corresponding annotator. All datasets are available online on Github\footnote{https://github.com/alan-turing-institute/TCPD}.

\subsubsection{Hyperparameters} The values for hyperparameters have been selected using a grid search over a limited number of values. We use the same hyperparameters for all datasets. The size of the sliding window $\mathit{w}$ is set to $6$ for all datasets. The number of hidden units of the LSTM network $U$ is set to 20. The number of the models in the TAEnet-ensemble $M$ is equal to $3$. The skip sizes $S$, for each TAEnet, have been set differently to $3$, $5$, and $7$, respectively, to learn data representation at various scales. The value for horizon hyperparameter $h$ is set to $4$. $C$ that determines the size of threshold (Equation \ref{eq:threshold_init}) is set to $1.4$; however, during the first few epochs after finding a change point, $C$ is set to a high value to let the network adapt itself to the new behavior. $n_{init}$ is set such that these initial samples do not contain any change points; in most experiments, we have selected $10\%$ of the samples for this purpose. Hyperparameter $\beta$ is set to $0.6$ so that the majority of the models should report a change point to make a change point decision. $n_{CPD}$ has been configured as $3$. When $n_{CPD}$ is set to a lower value, the model becomes highly sensitive to even minor data fluctuations, while a higher value may cause the model to overlook certain change points in the data. We train the network with Stochastic Gradient Decent (SGD) with a learning rate of $0.001$. The number of epochs $e_{init}$, $e_{train}$, and $e_{re-init}$ has been set to $10$, $5$, and $100$, respectively. The time series are standardized to have zero mean and unit variance. The results are an average of 10 random seeds.

\subsubsection{Implementation} We have implemented our proposed algorithm, ALACPD, in Python using Tensorflow \cite{tensorflow2015-whitepaper} and Keras \cite{chollet2015keras}. The start of our implementation are LSTNet\footnote{\url{https://github.com/fbadine/LSTNet}} \cite{lai2018modeling} and OED\footnote{\url{https://github.com/tungk/OED}} \cite{kieu2019outlier}.%

\begin{figure*}
     \centering
     
     \begin{subfigure}[b]{0.48\textwidth}
         \centering
         \includegraphics[width=\textwidth]{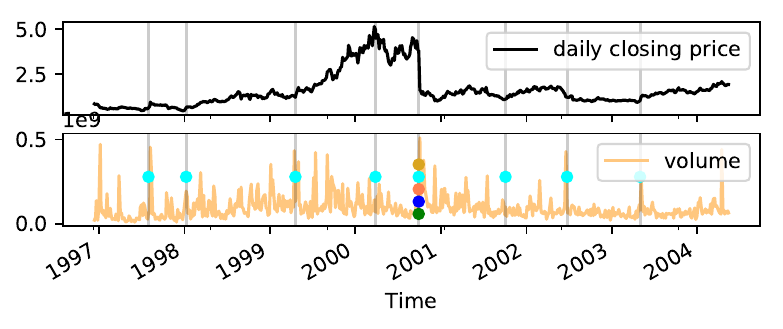}
         \caption{Apple}
         \label{fig:apple}
     \end{subfigure}
     \begin{subfigure}[b]{0.48\textwidth}
         \centering
         \includegraphics[width=\textwidth]{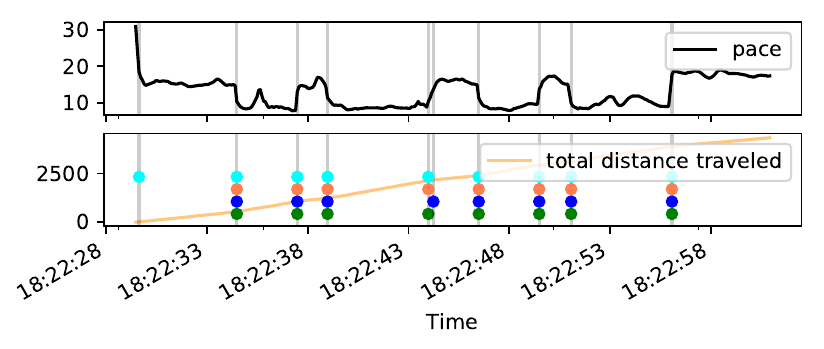}
         \caption{Run\_log}
         \label{fig:run_log}
     \end{subfigure}
        \
    \begin{subfigure}[b]{0.48\textwidth}
         \centering
         \includegraphics[width=\textwidth]{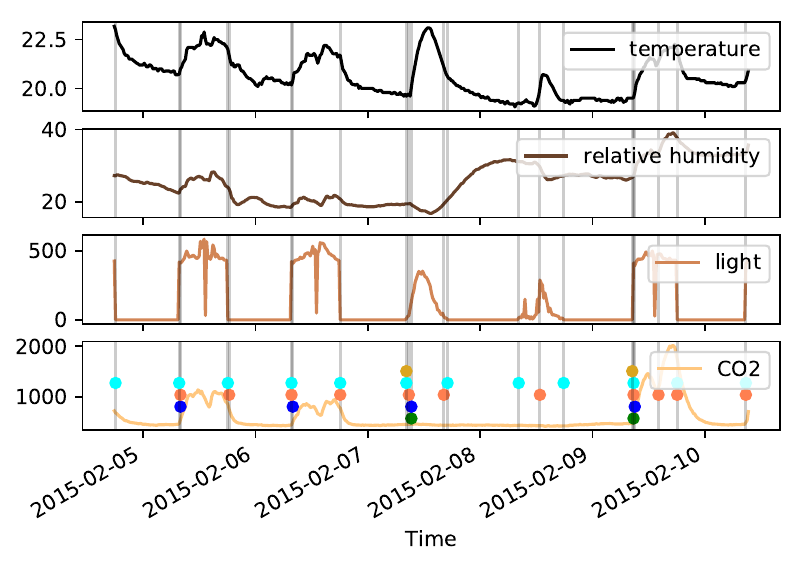}
         \caption{Occupancy}
         \label{fig:occupancy}
     \end{subfigure}
    \begin{subfigure}[b]{0.48\textwidth}
         \centering
         \includegraphics[width=\textwidth]{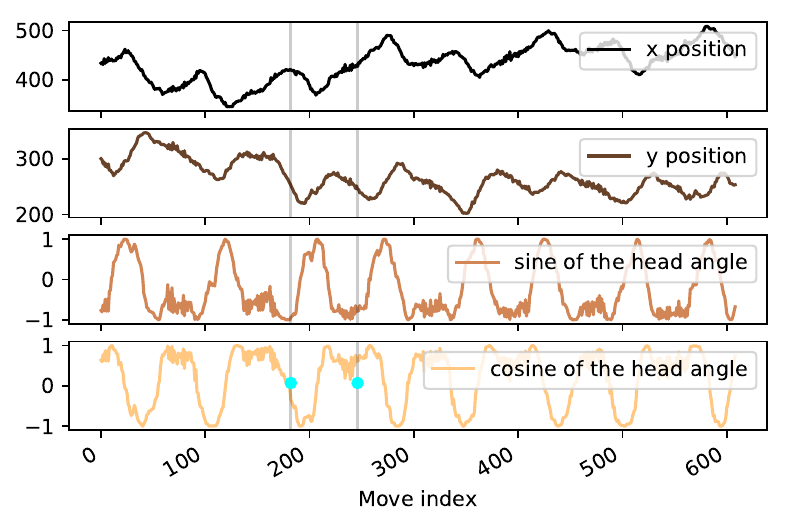}
         \caption{Bee\_waggle\_6}
         \label{fig:bee_waggle_6}
     \end{subfigure}
     \caption{Datasets and the change points annotations. The gray vertical lines depict all the change points determined by the annotator. The colored points on the lines refer to the annotators detected the corresponding change points; each color corresponds to a single annotator.}
        \label{fig:datasets}
\end{figure*}


\subsubsection{Baselines}\label{ssec:baselines} We compare ALACPD with several state-of-the-art unsupervised CPD methods. We mainly focus on the methods that can be applied to multivariate time series. The baselines that we have chosen for comparison include:
\begin{itemize}
    \item \textbf{BOCPD} \cite{adams2007bayesian} is an online Bayesian approach for CPD. It detects change points by estimating the posterior probability over the run length (the time since the last change point).
    \item \textbf{BOCPDMS} \cite{knoblauch2018spatio} is an extension of BOCPD to multiple models from a model universe $M$. 
    \item \textbf{ECP} \cite{matteson2014nonparametric} is an offline distance-based approach for CPD. This algorithm is nonparametric and uses pairwise energy distance among samples to perform CPD. However, it is computationally expensive when the dataset size is large.
    \item \textbf{KCPA} \cite{harchaoui2009kernel} is an offline kernel-based CPD algorithm. KCPA is also a nonparametric method that detects changes using a static test based on kernel Fisher discriminant ratio. This algorithm needs a maximum number of change points as input; in our experiments, we set it to $5$.
\end{itemize}
We have also considered the ZERO method as a baseline which always returns an empty set for the detected change points \cite{van2020evaluation}. The implementation of these methods is provided by TCPDBench\footnote{\url{https://github.com/alan-turing-institute/TCPDBench}} (Turing Change Point Detection Benchmark). 

It should be noted that we do not perform hyperparameter optimization for any of the algorithms. This is due to the fact that the focus of this paper is unsupervised CPD. In such settings in real-world applications, usually, ground truth for the change points is not available. Therefore, it is not possible to perform hyperparameter optimization in real-world applications. As a result, to get a realistic view of the performance of these algorithms, we use the best set of parameters used by the authors of each method in their papers. These hyperparameters have been discussed in \cite{van2020evaluation}.

\subsubsection{Evaluation Metrics}\label{ssec:metrics} In this paper, we use two evaluation metrics to measure the performance of each CPD method as introduced in \cite{van2020evaluation}. Evaluation metrics for CPD algorithms can be categorized into clustering and classification metrics. We have also chosen one metric from each category, as follows:

\begin{itemize}
    \item \textbf{Covering.} Covering is considered a clustering metric to evaluate CPD algorithms. This metric evaluates the method based on the quality of its segmentation. The Covering metric is formally defined as:
    \begin{equation}
        C(\mathcal{T}, \mathcal{G}_l)=  \sum_{\mathcal{A} \in \mathcal{G}_l} |\mathcal{A}|. \max_{\ \mathcal{A}' \in \mathcal{T}\ } J(\mathcal{A}, \mathcal{A}'),
    \end{equation}
    
    \begin{equation}
        J(\mathcal{A}, \mathcal{A}') = \frac{\mathcal{A} \cap \mathcal{A}'}{\mathcal{A} \cup \mathcal{A}'},
    \end{equation}
    where $C(\mathcal{T}, \mathcal{G}_l)$ is Covering score of the detected change points $\mathcal{T}$ and the ground-truth of change points' locations given by $l$-th annotator $\mathcal{G}_l$ ($l \in \{0, 1, ..., L\}$, $L$ is the number of annotators), and $J(\mathcal{A}, \mathcal{A}')$ is the Jaccard index of two sets $\mathcal{A}$ and $\mathcal{A}'$. As the annotation of the datasets is given by several experts, the mean of the Covering score for all of the annotators is reported as the final Covering score.
    
    \item \textbf{F1-score.} F1-score is a classification-based metric that measures the quality of the estimated change points. It is formally defined as:
    \begin{equation}\label{eq:fscore}
        F_1 = \frac{2PR}{P+R},
    \end{equation}
    
    \begin{equation}\label{eq:precision}
        P = \frac{|TP(\mathcal{G}, \mathcal{T})|}{|\mathcal{T}|},
    \end{equation}
    
    \begin{equation}\label{eq:recall}
        R = \frac{1}{L}\sum_{l=1}^{L}\frac{TP(\mathcal{G}_l, \mathcal{T})}{|\mathcal{G}_l|},
    \end{equation}

    where $F_1$ is F1-score (Equation \ref{eq:fscore}), $TP$ is the number of true positives, $P$ is called precision, $\mathcal{G}$ is the set of all human annotations, and $R$ is known as recall. F1-score incorporates both precision and recall to measure the effectiveness of an algorithm. \textbf{Precision} measures the relevancy of the detected change points; it is computed using the number of correctly identified change points ($TP$) over the number of detected change points. For computing $TP$ in Equation \ref{eq:precision}, the annotations for all the annotators are used. In addition, in evaluating change point detection algorithms with classification metrics, a margin of error of size $M$ is usually allowed for the distance of the detected change points and the true labels. \textbf{Recall} measures the effectiveness of the algorithm in finding true change points' locations; it is computed using the number of correctly identified change points ($TP$) over the number of true change points (Equation \ref{eq:recall}).
   
\end{itemize}


\renewrobustcmd{\bfseries}{\fontseries{b}\selectfont}
\renewrobustcmd{\boldmath}{}
\newrobustcmd{\bt}{\bfseries}
\begin{table}[!b]
\centering
    \caption{Performance comparison among all methods in terms of covering.}\label{tab:results_covering}
    \scalebox{0.8}{\parbox{\linewidth}{%
    \begin{tabular}{ccccccc|cc}\toprule 
        \bt Dataset        & \bt BOCPD &\bt  BOCPDMS&\bt ECP &\bt KCPA &\bt ZERO &\bt ALACPD &\bt \textbf{\begin{tabular}[c]{@{}c@{}}ALACPD\\ w/oAR\end{tabular}}  &\bt \textbf{\begin{tabular}[c]{@{}c@{}}ALACPD\\ w/oAE\end{tabular}} \\\midrule
        \bt Apple          & 0.401 & 0.334 & 0.313 & 0.461 & 0.424 & \bt0.513 $\pm$ 0.01 & 0.421 $\pm$ 0.00& 0.533 $\pm$ 0.01\\
        \bt Occupancy      & 0.549 & 0.458 & 0.519 & 0.575 & 0.235 & \bt0.599 $\pm$ 0.03& 0.570 $\pm$ 0.02& 0.616 $\pm$ 0.02\\
        \bt Run\_log       & \bt0.815 & 0.301 & 0.630 & 0.631 & 0.303  & 0.708 $\pm$ 0.03 & 0.695 $\pm$ 0.01& 0.592 $\pm$ 0.09\\
        \bt Bee\_waggle\_6 & 0.089 & 0.887 & 0.116 & 0.653 &\bt 0.891  & 0.256 $\pm$ 0.04& 0.290 $\pm$ 0.03& 0.315 $\pm$ 0.08\\\bottomrule         
    \end{tabular}
    }}

\end{table}

\begin{table}[!b]
\centering
    \caption{Performance comparison among all methods in terms of F1-score.}\label{tab:results_f_measure}
    \scalebox{0.8}{\parbox{\linewidth}{%
    \begin{tabular}{ccccccc|cc}\toprule 
    \bt Dataset        & \bt BOCPD  &\bt BOCPDMS &\bt ECP &\bt KCPA &\bt ZERO &\bt ALACPD &\bt \textbf{\begin{tabular}[c]{@{}c@{}}ALACPD\\ w/oAR\end{tabular}}  &\bt \textbf{\begin{tabular}[c]{@{}c@{}}ALACPD\\ w/oAE\end{tabular}} \\\midrule
    \bt Apple          & 0.606 & 0.381& 0.513 & 0.573 & 0.593 & \bt0.761 $\pm$ 0.05 & 0.403 $\pm$ 0.07 & 0.821 $\pm$ 0.05\\
    \bt Occupancy      & \bt0.807 & 0.496 & 0.716 & 0.619 & 0.340 & 0.797 $\pm$ 0.03& 0.565 $\pm$ 0.03& 0.796 $\pm$ 0.07\\
    \bt Run\_log       & \bt1.000 & 0.420 & 0.749 & 0.782 & 0.445  & 0.848 $\pm$ 0.07 & 0.719 $\pm$ 0.06& 0.803 $\pm$ 0.09\\
    \bt Bee\_waggle\_6 & 0.120 & 0.481 & 0.116 & 0.437 & \bt0.928  & 0.256 $\pm$ 0.05& 0.366 $\pm$ 0.05& 0.250 $\pm$ 0.06\\\bottomrule       
    \end{tabular}
    }}
\end{table}

\subsection{Results}\label{ssec:results}
In this section, we summarize the results of the experiments. We perform CPD using ALACPD and the methods described in Section \ref{ssec:baselines} on four real-world time series introduced in Section \ref{ssec:datasets}. We evaluate all the methods using the two evaluation metrics explained in Section \ref{ssec:metrics}. The results are summarized in Tables \ref{tab:results_covering}, and \ref{tab:results_f_measure}. The last two columns of these tables present the results of an ablation study which will be discussed in Section \ref{ssec:ablation}. The locations of the detected change points for each method and dataset are presented in Appendix \ref{appendix:results_visualization}.

As can be seen in Tables \ref{tab:results_covering} and \ref{tab:results_f_measure}, ALACPD and BOCPD are the best performers in terms of both Covering and F1-score on Apple and run\_log datasets, respectively. As can be seen in Figure \ref{fig:datasets}, Apple and run\_log datasets have different types of changes. While the run\_log dataset has clear scale changes in its time series that the annotators unanimously detected those points, the changes in the Apple dataset are more difficult to detect for most of the methods, and even for human annotators. While the majority of the methods fail to accurately detect changes on the Apple dataset, ALACPD has decent performance on this dataset. It outperforms the second-best performer in terms of Covering by $0.052$ and in terms of F1-score by $0.155$ gap. BOCPDMS has the worst performance among methods on these datasets; it has been even outperformed by the ZERO method in terms of both evaluation metrics. Moreover, the two offline algorithms, ECP and KCPA, have not been able to outperform online models on these two datasets. ALACPD outperforms both methods in terms of Covering and F1-score in most cases considered.

On the occupancy dataset, ALACPD and BOCPD are the best performers in terms of Covering and F1-score, respectively. As shown in Figure \ref{fig:cpd_occupancy} in Appendix \ref{appendix:results_visualization}, on this dataset BOCPD detects the maximum number of change points among methods, and it has many false positives. The other considered methods have been outperformed by KCPA on this dataset. 
As this dataset contains several change points, the ZERO method, which returns no change points in all cases, is the worst performer on this dataset.   
On the other hand, ZERO outperforms all the methods on the Bee\_waggle\_6 dataset, which contains only $2$ change point labels. BOCPD and ECP have the worst performance on the Bee\_waggle\_6 dataset. This shows that these methods are extremely sensitive to the scale of the input data. ALACPD is less sensitive to local changes than BOCPD and ECP; however, it is outperformed by BOCPDMS and KCPA on the Bee\_waggle\_6 dataset. It should be noted that KCPA needs the maximum number of change points in the series as a hyperparameter. That is the reason why it detects a low number of change points on the Bee\_waggle\_6 dataset. 

\begin{figure}[!t]
     \centering
     
     \begin{subfigure}[b]{0.48\columnwidth}
         \centering
         \includegraphics[width=0.9\textwidth]{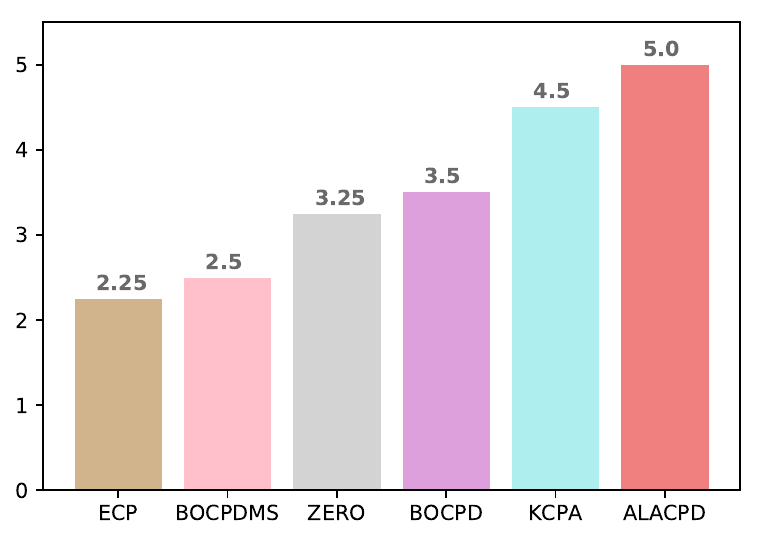}
         \caption{Covering}
         \label{fig:covering}
     \end{subfigure}
     \begin{subfigure}[b]{0.48\columnwidth}
         \centering
         \includegraphics[width=0.9\textwidth]{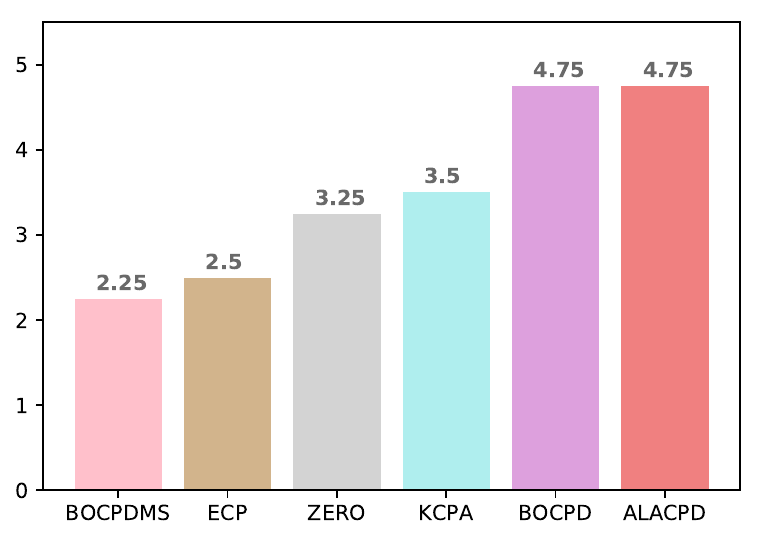}
         \caption{F1-score}
         \label{fig:fscore}
     \end{subfigure}
        \
    
     \caption{Comparison of average ranking scores. The scores for each method are an average of the ranking of that method over various datasets in terms of the corresponding evaluation metric.}
        \label{fig:evaluation}
\end{figure}

\begin{table}[!b]
\centering\centering
\crefformat{footnote}{#2\footnotemark[#1]#3}
    \caption{Summary Table of CPD algorithms}\label{tab:comparison}
    \scalebox{0.8}{\parbox{\linewidth}{%
    \begin{tabular}{c@{\hskip 0.05in}c@{\hskip 0.05in}c@{\hskip 0.05in}c@{\hskip 0.05in}c@{\hskip 0.05in}c@{\hskip 0.05in}c@{\hskip 0.05in}c@{\hskip 0.05in}c}\toprule 
        \bt Dataset        & \bt Covering\footnote{\label{fn:check}\scriptsize{The check-marks indicate the relative performance of methods in terms of the ranking in the corresponding metric in Figure \ref{fig:evaluation}; one check-mark indicates that the corresponding method is among the worst performers in the corresponding metric.}} & \bt F1-score\textsuperscript{\ref{fn:check} }& \bt Processing Delay  &\bt \makecell{Algorithm\\Assumptions}  &\bt \makecell{Computational \\complexity w.r.t $n$} & \bt Additional Remarks\\\midrule
        \bt BOCPD   &\checkmark\checkmark & \checkmark\checkmark\checkmark & Online    & Parametric    & $O(r(t))$\footnote{\label{fn:cost}\scriptsize{The computational complexity of this method can be reduced by using various techniques, e.g., pruning the run length distribution \cite{knoblauch2018spatio}.} }  & \makecell{ $r(t)$ refers to run length which is the\\ distance since the previous change point.}\\\hline
        \bt BOCPDMS &\checkmark & \checkmark &Online     & Parametric    & $O(|\mathcal{M}|r(t))$\textsuperscript{\ref{fn:cost}} & \makecell{$|\mathcal{M}|$ is the number of models in $M$}\\\hline
        \bt ECP     &\checkmark & \checkmark&Offline    & Nonparametric & $O(n^2)$ &-\\\hline
        \bt KCPA    & \checkmark\checkmark\checkmark& \checkmark\checkmark&Offline    & Nonparametric &$O(K_{max}n^2)$ & \makecell{$K_{max}$ is the maximum number of\\ change points in the time series.}\\\hline
        \bt ALACPD  & \checkmark\checkmark\checkmark& \checkmark\checkmark\checkmark&Online     & Parametric    &$O(w)$&- \\\bottomrule       
    \end{tabular}
    }}

\end{table}

To summarize the results, we have computed the average ranking of all methods for different datasets in terms of the Covering and F1-score metrics. The results are presented in Figure \ref{fig:evaluation}. In this figure, ALACPD is the best performer among methods in terms of the Covering score, and it is on par with the best performer in terms of the F1-score among the considered unsupervised CPD methods. BOCPD, the major competitor of ALACPD, has been outperformed by ALACPD with a significant gap in the Covering score. This is probably because BOCPD, with its default parameters, is very sensitive to abrupt changes in the scale of the data. As a result, BOCPD performs decently on the Run\_log and Occupancy datasets that contain such changes, while it fails to perform well on the Apple and Bee\_waggle\_6 datasets where the input scale changes continuously. 

Overall, it can be observed in Figure \ref{fig:evaluation} that while there are some subtle differences in the average ranking of the methods in terms of Covering and F1-score, there is a consistency in the overall ranking. ALACPD, BOCPD, and KCPA are among the top three performers in both metrics. On the other hand, ECP and BOCPDMS have been outperformed by the ZERO method in both cases. Therefore, it can be concluded that the ranking of the methods is not highly dependent on the choice of the evaluation metric.


\section{Discussion}
Throughout this section, we provide the results of three studies, including a performance evaluation of ALACPD and detailed comparison with the considered CPD methods, CPD in human activity dataset using ALACPD, and an ablation study to measure the effectiveness of the sub-components of ALACPD. 
\subsection{Performance Evaluation}\label{ssec:discussion_performance}
In this section, we provide an overview of the performance of ALACPD and compare it with the other considered CPD methods in terms of the quality of the CPD, processing delay, algorithm assumptions, and computational complexity. The results are summarized in Table \ref{tab:comparison}.

As shown in Table \ref{tab:comparison}, ALACPD is among the best two performers in terms of both Covering and F1-score. It processes the data online using a sliding window over the time series. It discards old samples after adapting the model's parameters. Therefore, the time and space complexity of the model is equal to $O(w)$. BOCPD (the main competitor of ALACPD in terms of F1-score) and BOCPDMS have a higher computational complexity than ALACPD. The complexity of the original methods is in order of $r(t)$, where $r(t)$ is the distance from the previous change point. Therefore, a rise in $r(t)$ adversely affects the complexity of this algorithm. However, various techniques have been proposed to address this complexity \cite{knoblauch2018spatio}. Finally, the complexity of the two offline methods is in the order of $n^2$, ECP, and KCPA, which are significantly higher than the other ones. These methods are impractical when the time series length $n$ (number of total observations) is large. 

In short, ALACPD ensures decent CPD performance while maintaining a low complexity w.r.t $n$. While ALACPD is a parametric model, we showed that the default values for hyperparameters work well among different datasets.

\subsection{Change Point Detection for Human Activity Recognition}
In this section, we analyze the performance of ALACPD in detecting changes in human activities. For this purpose, we use the Human Activity Recognition (HAR) dataset from the UCI data repository \footnote{\url{https://archive.ics.uci.edu/}} \cite{anguita2013public}. This dataset is the observations of $30$ subjects performing $6$ activities, including walking, standing, laying, and sitting. The data was recorded by a smartphone connected to the subjects' bodies. This time series has 561 dimensions and 7352 training samples. Therefore, it is a good benchmark to measure the performance of the algorithms on complex multi-dimensional changes.

\begin{table}[!t]
    \centering
    \caption{Performance comparison between ALACPD and BOCPD.} \label{tab:har_results}
    \scalebox{0.85}{\parbox{\linewidth}{%
    \begin{tabular}{c@{\hskip 0.05in}c@{\hskip 0.05in}c@{\hskip 0.05in}c@{\hskip 0.05in}c@{\hskip 0.05in}c@{\hskip 0.05in}c@{\hskip 0.05in}c@{\hskip 0.05in}c@{\hskip 0.05in}c@{\hskip 0.05in}c@{\hskip 0.05in}c@{\hskip 0.05in}c@{\hskip 0.05in}c@{\hskip 0.05in}c@{\hskip 0.05in}c@{\hskip 0.05in}c}
        \toprule
        & & \multicolumn{15}{c}{\pmb{ $F1-score$}}\\
        \cmidrule(l){3-17}
         & &\multicolumn{3}{c}{\pmb{ $M=2$}} & \multicolumn{3}{c}{\pmb{ $M=3$}} & \multicolumn{3}{c}{\pmb{ $M=4$}} & \multicolumn{3}{c}{\pmb{ $M=5$}} & \multicolumn{3}{c}{\pmb{ $M=6$}} \\ \cmidrule(l){3-5}\cmidrule(l){6-8}\cmidrule(l){9-11} \cmidrule(l){12-14}\cmidrule(l){15-17}
         \bt Method & \pmb{$Covering$} &\bt F\textsubscript{1} &\bt P & \bt R   &\bt F\textsubscript{1}  &\bt P & \bt R      &\bt F\textsubscript{1}  &\bt P & \bt R      &\bt F\textsubscript{1}  &\bt P & \bt R     &\bt F\textsubscript{1}  &\bt P & \bt R\\ \midrule
            ALACPD & \pmb{$0.628$}&\pmb{$0.673$}&$0.71$&\pmb{$0.639$}&\pmb{$0.68$}&$0.718$&\pmb{$0.646$}&\pmb{$0.695$}&$0.734$&\pmb{$0.661$}&\pmb{$0.707$}&$0.746$&\pmb{$0.671$}&\pmb{$0.722$}&$0.762$&\pmb{$0.686$}\\
            BOCPD & $0.596$ &$0.63$&\pmb{$0.726$}&$0.557$&$0.675$&\pmb{$0.777$}&$0.596$&$0.679$&\pmb{$0.781$}&$0.6$&$0.695$&\pmb{$0.8$}&$0.614$&$0.707$&\pmb{$0.814$}&$0.625$\\

        \bottomrule
    \end{tabular}  
    }}
\end{table}

We perform unsupervised CPD with ALACPD on the HAR dataset and measure its performance in detecting changes between various activities. The change point labels are the time steps where we observe a change in the activity. We compare the results with its main competitor, BOCPD\footnote{We have also tried to run BOCPDMS on this dataset; however, the running time exceeded our considered time limit (6 hours), and we were not able to get the results.}. The results are depicted in Table \ref{tab:har_results}. On the HAR dataset, ALACPD has a higher Covering and F1-score (for various error margins $M$) than BOCPD. As seen in the F1-score results, BOCPD recalls fewer true change points than ALACPD while being more precise in detecting those locations. Overall, the F1-score, which shows the trade-off between precision and recall, for ALACPD is higher than BOCPD.

\begin{table}[!b]
\centering
    \caption{Human Activity Recognition Analysis}\label{tab:activity}
    \scalebox{0.9}{\parbox{\linewidth}{%
    \begin{tabular}{ccccccccc}\toprule 
        \bt Index        &\bt Activity Change & \bt \# of samples  &\bt Recall\textsubscript{ALACPD}   &\bt Recall\textsubscript{BOCPD}\\\midrule
        \bt 1   & Walking \textrightarrow Walking downstairs & 42 & \bt 0.81  & 0.69\\ 
        \bt 2   & Walking upstairs \textrightarrow Walking downstairs & 14 & \bt  0.93 &0.71\\ 
        \bt 3   & Walking upstairs \textrightarrow Standing & 39 & 0.90 &  \bt 1\\ 
        \bt 4   &  Walking downstairs \textrightarrow  Walking upstairs  & 54& 0.31 & \bt 0.76\\ 
        \bt 5   & Walking downstairs \textrightarrow Standing & 2 & \bt 1 & \bt 1\\ 
        \bt 6   & Sitting \textrightarrow Laying & 43& \bt 0.67 &0.09\\ 
        \bt 7   &  Standing \textrightarrow Sitting & 42 & \bt 0.42 &0.1\\ 
        \bt 8   &  Laying \textrightarrow Walking & 42 &0.93 & \bt 1\\ 
        \bt 9   & Laying \textrightarrow Sitting & 1&0 &0\\ \midrule 
        \bt 10   & All & 279& \bt  0.67 &0.61 \\ 
        \bottomrule       
    \end{tabular}
    }}
\end{table}


Next, we elaborate on the detected change points by ALACPD on the HAR dataset. We compute the recall ($M=5$) for each type of change point separately. By type of change point, we refer to the type of activity changes. These changes in the HAR dataset, along with their number and corresponding recall for ALACPD are summarized in Table \ref{tab:activity}. From Table \ref{tab:activity}, it can be observed that ALACPD is good at finding the changes related to walking except walking from downstairs to upstairs ($R=0.31$). Besides, detecting changes related to sitting is also difficult for ALACPD. While BOCPD fails at finding changes for the sitting-related activities (activity changes $6$ and $7$) with $ recall <0.1$, ALACPD achieves $0.67$ and $0.42$ on these activities, respectively.

\subsection{Ablation Study}\label{ssec:ablation}
This section describes the results of an ablation study on ALACPD. In this study, we analyze the effectiveness of recurrent and AR components in the performance of ALACPD. We consider two variants of ALACPD:

\begin{enumerate}
    \item \textbf{\footnotesize{ALACPDw/oAR}} exploits only the recurrent unit in the TAEnet architecture. 
    \item \textbf{\footnotesize{ALACPDw/oAE}} only has the AR component in the TAEnet architecture.
\end{enumerate}

We repeat the experiments from Section \ref{ssec:results}. The results for ALACPDw/oAR and ALACPDw/oAE are presented in Tables \ref{tab:results_covering} and \ref{tab:results_f_measure}. The locations of the detected change points for these methods are presented in Appendix \ref{appendix:results_visualization}.

As can be seen from Figures in Appendix \ref{appendix:results_visualization}, ALACPDw/oAR has a high sensitivity in CPD, particularly on the Apple dataset. ALACPDw/oAE, in most cases, quickly responds to sudden scale changes and detects these change points. However, when the changes take several steps to happen, it might fail to detect such changes, e.g., the change happening around 18:22:38 on the run\_log dataset or some change points on the occupancy dataset. In such cases, the AR component learns this behavior smoothly and adapts its parameters gradually to this new data distribution. We have already partly alleviated this problem by using the hyperparameter $h$. we have used the horizon $h$ parameter to forecast the current sample using the samples from $h$ steps earlier. Therefore, this hinders the AR component from quickly adapting the parameters to the new data distribution. While this solution helps ALACPDw/oAE detect some change points, it does not completely address this problem.

ALACPD aims to exploit both recurrent and AR components to learn the data representation effectively. As explained in Section \ref{sssec:TAEnet}, TAEnet learns an effective combination of these two components during the training. As a result, it outperforms ALACPDw/oAR and ALACPDw/oAE in most cases considered in terms of the Covering and F1-score.

\section{Conclusion}
\looseness=-1
In this paper, we presented a new deep learning approach for the problem of change point detection, named ALACPD. By using an LSTM-Autoencoder to learn long-term dependencies, in combination with an Auto-regressive model to respond to local scale changes in the data rapidly, our model is able to detect changes in an unsupervised online memory-free manner. Our findings demonstrate that ALACPD is the best performer among other considered state-of-the-art CPD algorithms in terms of the Covering metric, and it is on par with the best performer in terms of the F1-score in unsupervised online CPD from several real-world multi-dimensional time series. To further our research, we intend to extend our method to perform hierarchical CPD to detect changes over long periods. 

\textbf{Data availability.} All datasets used in this study are available online on \url{https://github.com/alan-turing-institute/TCPD}.

\section*{Compliance with ethical standards}

\textbf{Conflict of interest} The authors declare that they have no conflict of
interest.

\bibliographystyle{unsrt}
\bibliography{resources}

\newpage

\begin{appendices}

\section{Results Visualization}\label{appendix:results_visualization}
\begin{figure*}[!ht]
     \centering
     
     \begin{subfigure}[b]{0.48\textwidth}
         \centering
         \includegraphics[width=\textwidth]{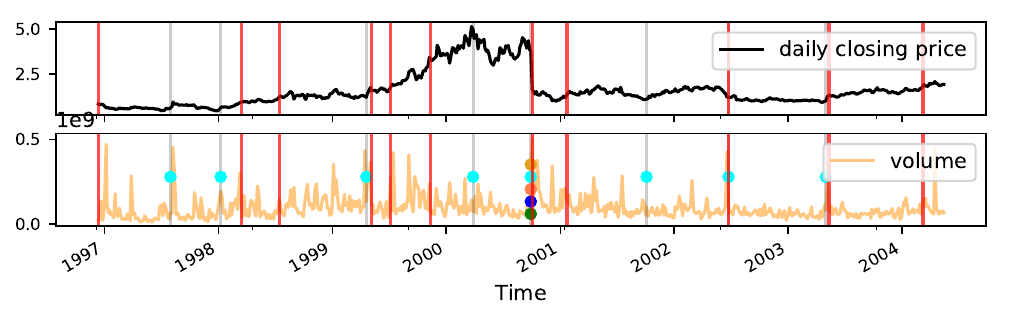}
         \vspace*{-6mm}
         \caption{BOCPD}
         \label{fig:apple_bocpd}
     \end{subfigure}
     \begin{subfigure}[b]{0.48\textwidth}
         \centering
         \includegraphics[width=\textwidth]{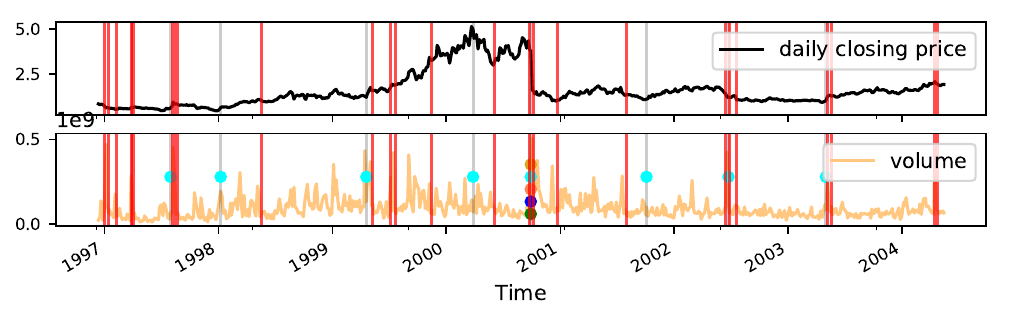}
         \vspace*{-6mm}
         \caption{BOCPDMS}
         \label{fig:apple_bocpdms}
     \end{subfigure}
        \
    \begin{subfigure}[b]{0.48\textwidth}
         \centering
         \includegraphics[width=\textwidth]{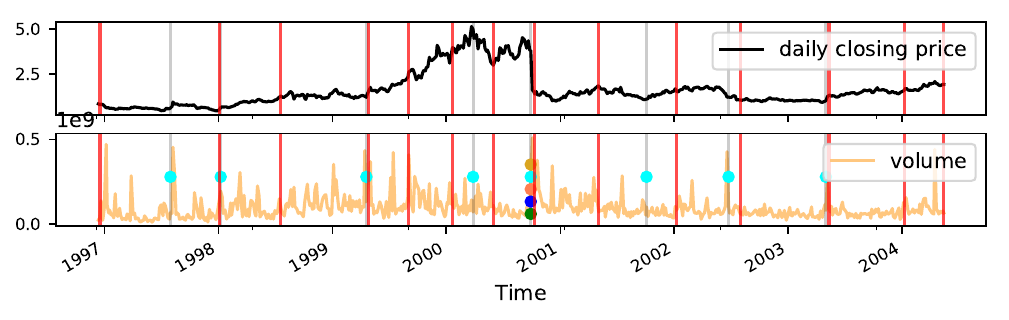}
         \vspace*{-6mm}
         \caption{ECP}
         \label{fig:apple_ecp}
     \end{subfigure}
    \begin{subfigure}[b]{0.48\textwidth}
         \centering
         \includegraphics[width=\textwidth]{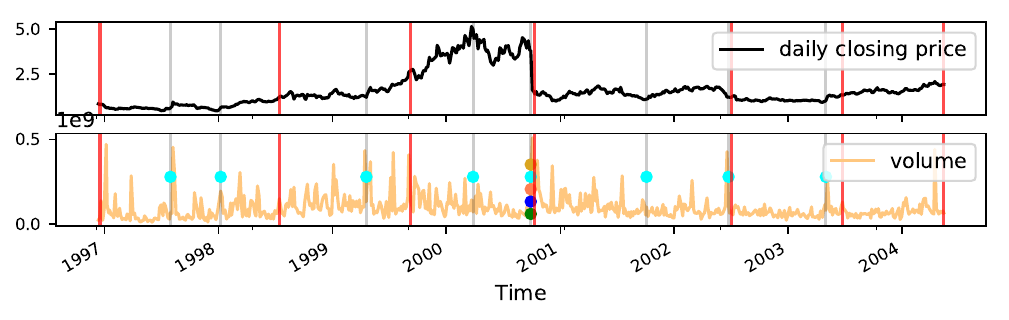}
         \vspace*{-6mm}
         \caption{KCPA}
         \label{fig:apple_kcpa}
     \end{subfigure}
     \begin{subfigure}[b]{0.48\textwidth}
         \centering
         \includegraphics[width=\textwidth]{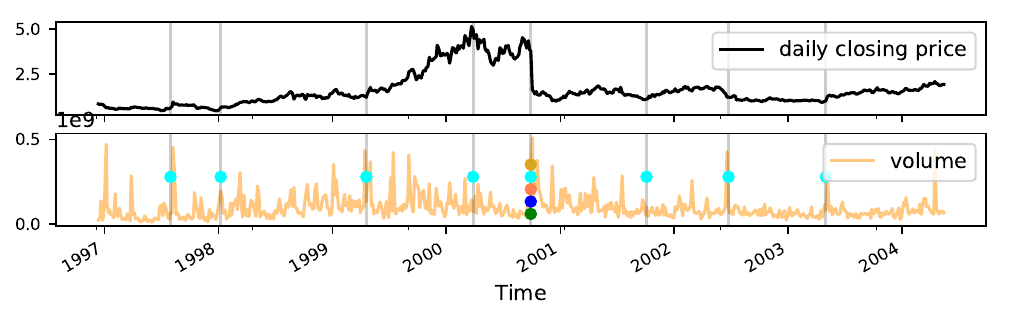}
         \vspace*{-6mm}
         \caption{ZERO}
         \label{fig:apple_zero}
     \end{subfigure}
     \begin{subfigure}[b]{0.48\textwidth}
         \centering
         \includegraphics[width=\textwidth]{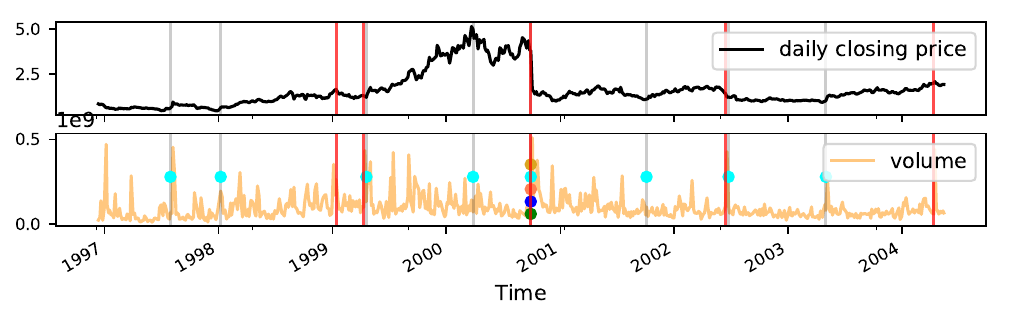}
         \vspace*{-6mm}
         \caption{ALACPD}
         \label{fig:apple_alacpd}
     \end{subfigure}
     \begin{subfigure}[b]{0.48\textwidth}
         \centering
         \includegraphics[width=\textwidth]{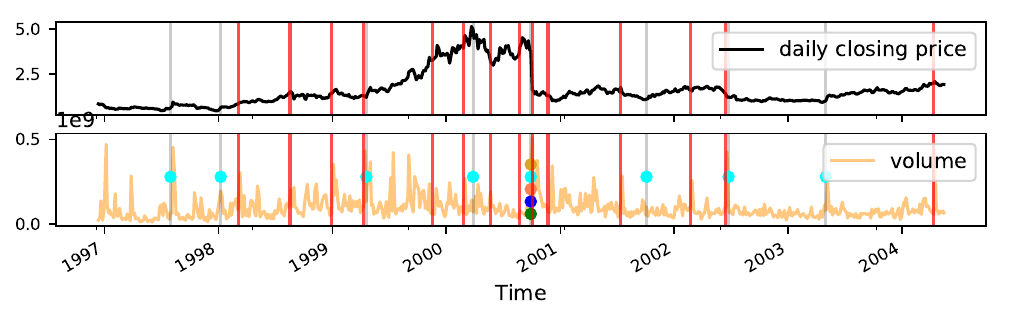}
         \vspace*{-6mm}
         \caption{ALACPDw/oAR}
         \label{fig:apple_alacpd_lstm}
     \end{subfigure}
     \begin{subfigure}[b]{0.48\textwidth}
         \centering
         \includegraphics[width=\textwidth]{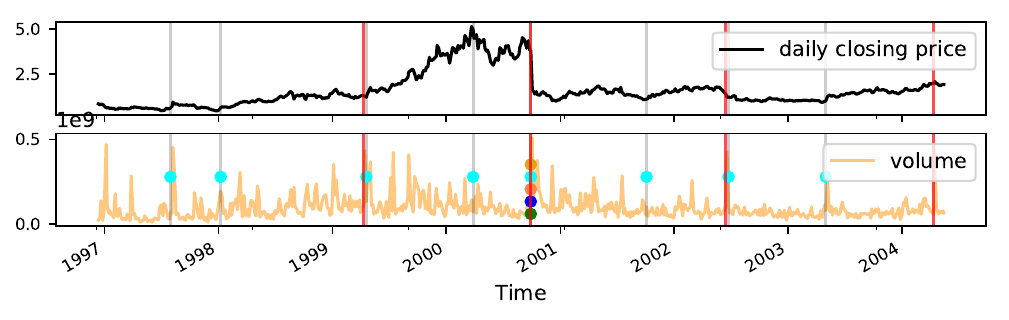}
         \vspace*{-6mm}
         \caption{ALACPDw/oAE}
         \label{fig:apple_alacpd_ar}
     \end{subfigure}

     \caption{CPD results on the Apple dataset. The gray and red vertical lines depict all the change points determined by the annotators and the algorithm, respectively. The colored points on the lines refer to the annotators detected the corresponding change points; each color corresponds to a single annotator.}
        \label{fig:cpd_apple}
\end{figure*}

\begin{figure*}[!ht]
     \centering
     
     \begin{subfigure}[b]{0.48\textwidth}
         \centering
         \includegraphics[width=\textwidth]{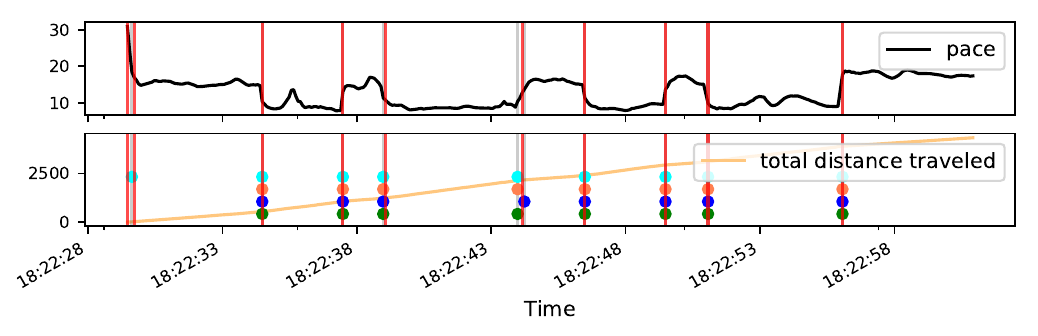}
         \caption{BOCPD}
         \label{fig:run_log_bocpd}
     \end{subfigure}
     \begin{subfigure}[b]{0.48\textwidth}
         \centering
         \includegraphics[width=\textwidth]{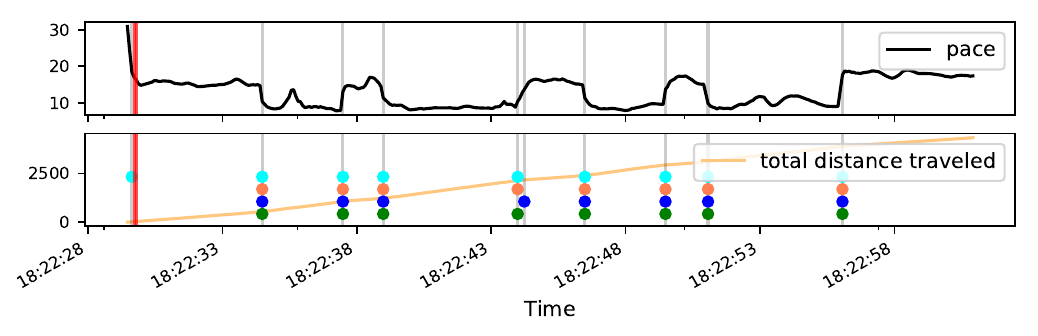}
         \caption{BOCPDMS}
         \label{fig:run_log_bocpdms}
     \end{subfigure}
        \
    \begin{subfigure}[b]{0.48\textwidth}
         \centering
         \includegraphics[width=\textwidth]{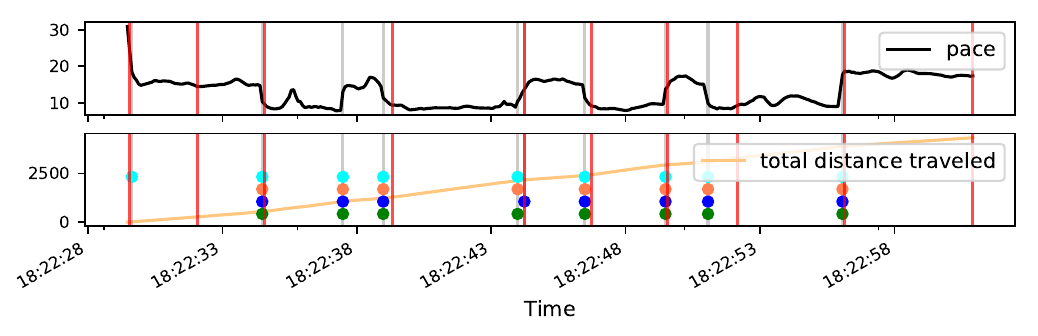}
         \caption{ECP}
         \label{fig:run_log_ecp}
     \end{subfigure}
    \begin{subfigure}[b]{0.48\textwidth}
         \centering
         \includegraphics[width=\textwidth]{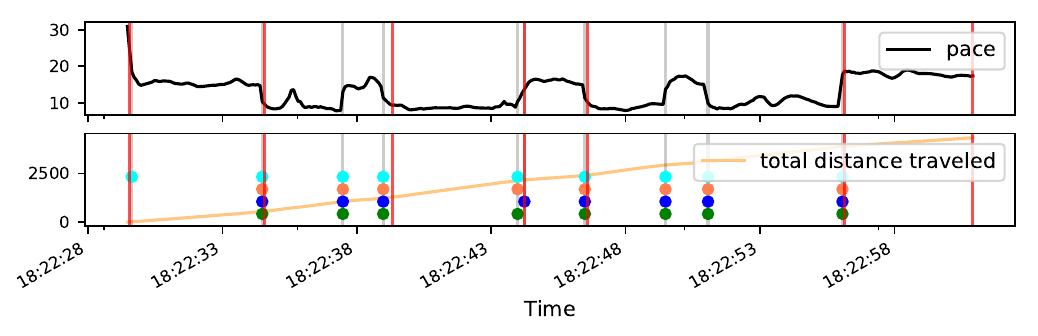}
         \caption{KCPA}
         \label{fig:run_log_kcpa}
     \end{subfigure}
     \begin{subfigure}[b]{0.48\textwidth}
         \centering
         \includegraphics[width=\textwidth]{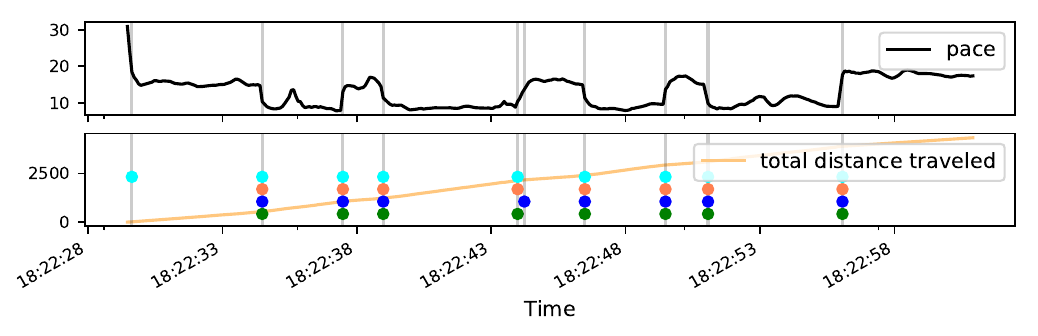}
         \caption{ZERO}
         \label{fig:run_log_zero}
     \end{subfigure}
      \begin{subfigure}[b]{0.48\textwidth}
         \centering
         \includegraphics[width=\textwidth]{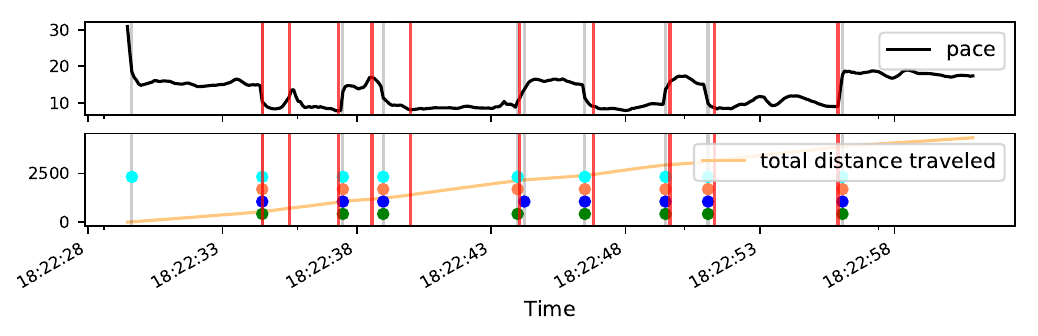}
         \caption{ALACPD}
         \label{fig:run_log_alacpd}
     \end{subfigure}
     \begin{subfigure}[b]{0.48\textwidth}
         \centering
         \includegraphics[width=\textwidth]{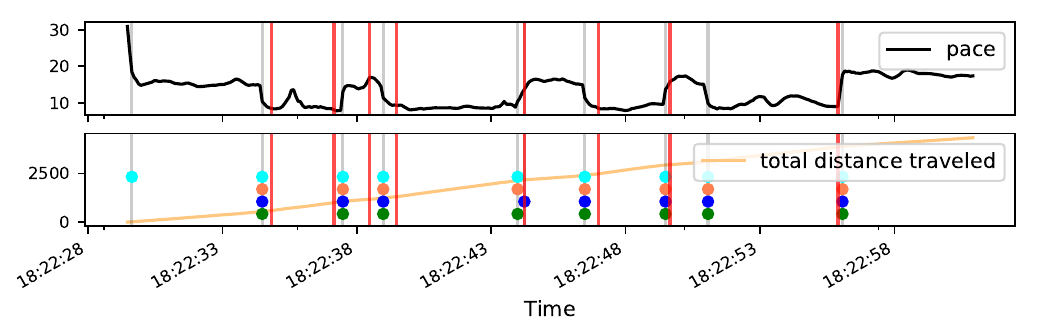}
         \caption{ALACPDw/oAR}
         \label{fig:run_log_alacpd_lstm}
     \end{subfigure}
     \begin{subfigure}[b]{0.48\textwidth}
         \centering
         \includegraphics[width=\textwidth]{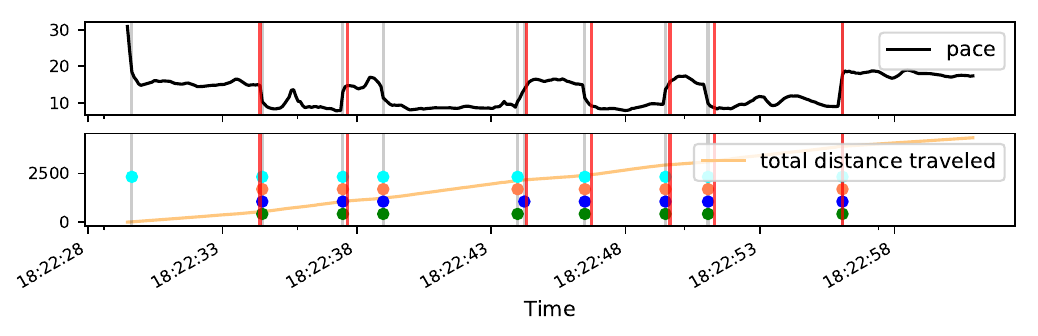}
         \caption{ALACPDw/oAE}
         \label{fig:run_log_alacpd_ar}
     \end{subfigure}
     
     \caption{CPD results on the Run\_log dataset. The gray and red vertical lines depict all the change points determined by the annotators and the algorithm, respectively. The colored points on the lines refer to the annotators detected the corresponding change points; each color corresponds to a single annotator.}
        \label{fig:cpd_run_log}
\end{figure*}

\begin{figure*}[!ht]
     \centering
     
     \begin{subfigure}[b]{0.48\textwidth}
         \centering
         \includegraphics[width=\textwidth]{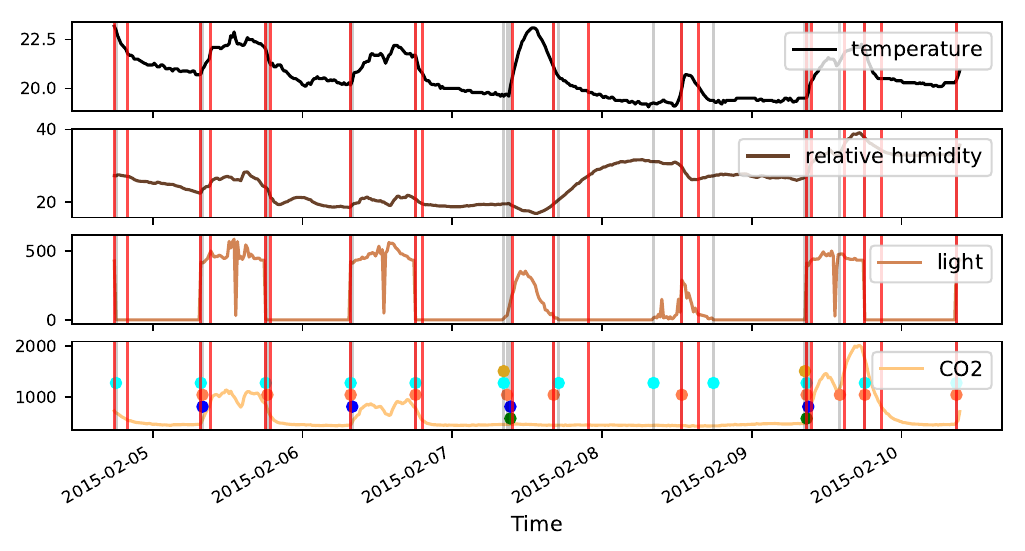}
         \vspace*{-6mm}
         \caption{BOCPD}
         \label{fig:occupancy_bocpd}
     \end{subfigure}
     \begin{subfigure}[b]{0.48\textwidth}
         \centering
         \includegraphics[width=\textwidth]{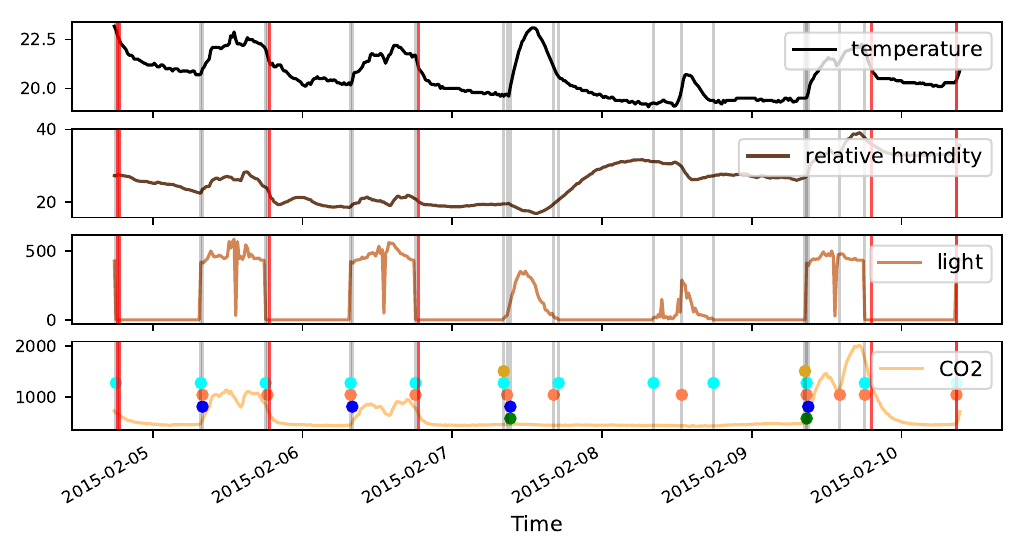}
         \vspace*{-6mm}
         \caption{BOCPDMS}
         \label{fig:occupancy_bocpdms}
     \end{subfigure}
        \
    \begin{subfigure}[b]{0.48\textwidth}
         \centering
         \includegraphics[width=\textwidth]{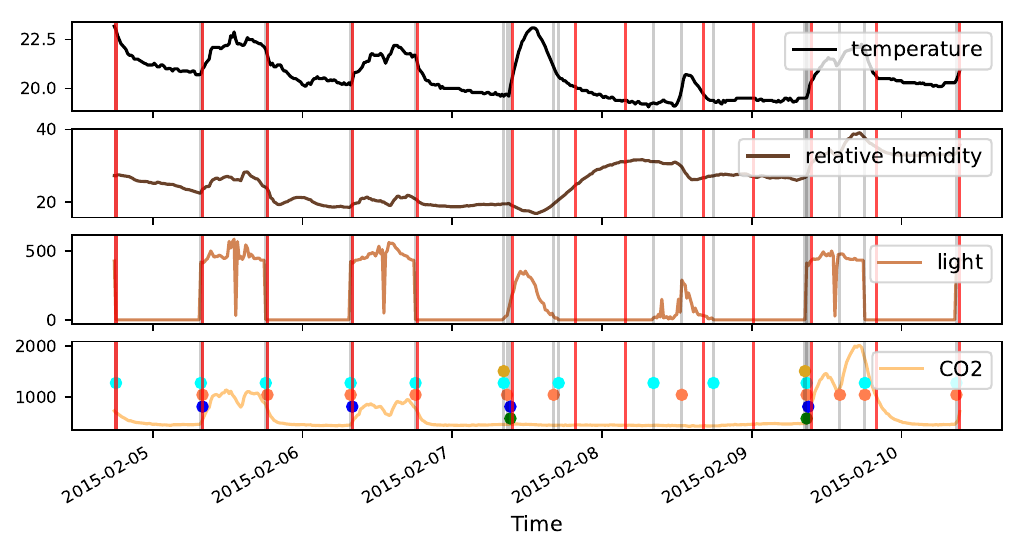}
         \vspace*{-6mm}
         \caption{ECP}
         \label{fig:occupancy_ecp}
     \end{subfigure}
    \begin{subfigure}[b]{0.48\textwidth}
         \centering
         \includegraphics[width=\textwidth]{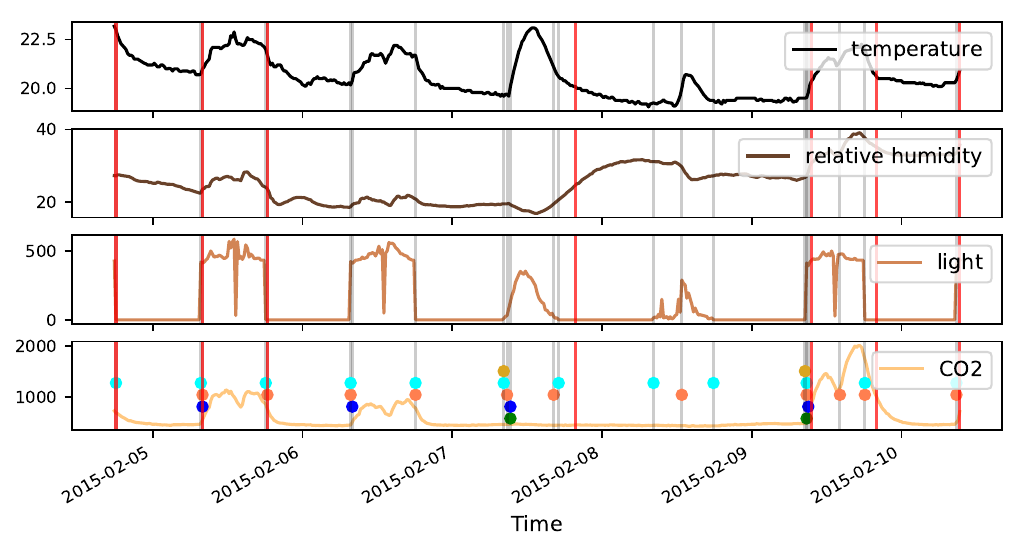}
         \vspace*{-6mm}
         \caption{KCPA}
         \label{fig:occupancy_kcpa}
     \end{subfigure}
     \begin{subfigure}[b]{0.48\textwidth}
         \centering
         \includegraphics[width=\textwidth]{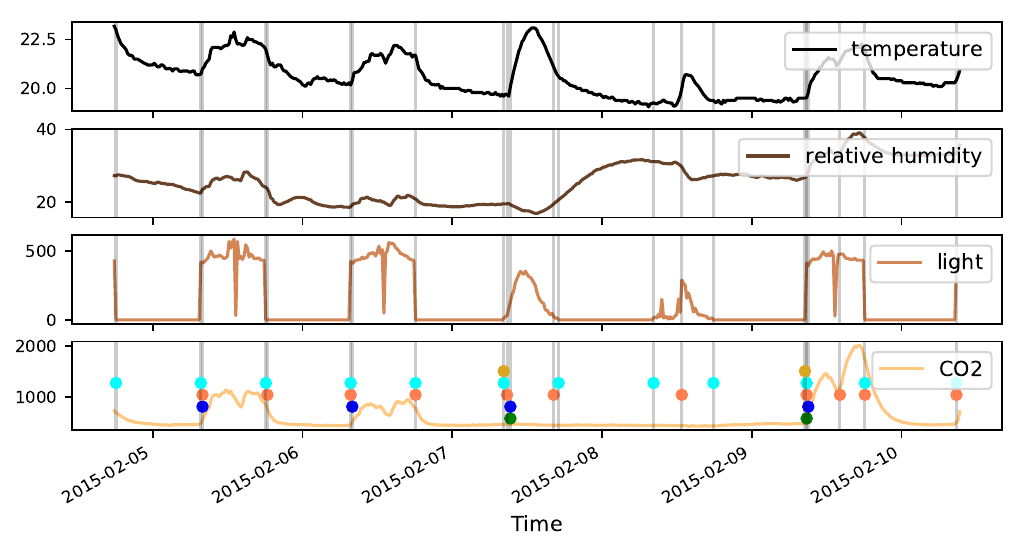}
         \vspace*{-6mm}
         \caption{ZERO}
         \label{fig:occupancy_zero}
     \end{subfigure}
     \begin{subfigure}[b]{0.48\textwidth}
         \centering
         \includegraphics[width=\textwidth]{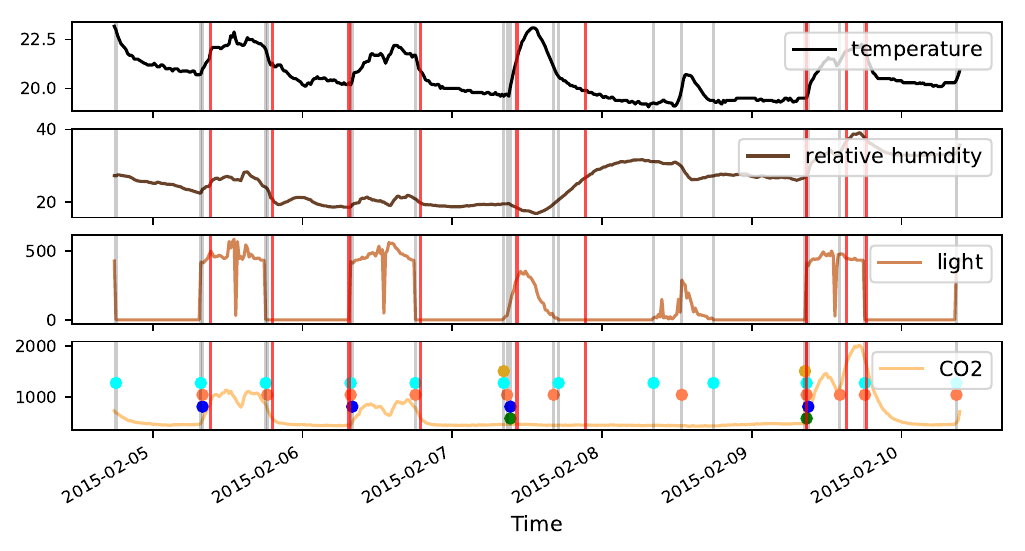}
         \vspace*{-6mm}
         \caption{ALACPD}
         \label{fig:occupancy_alacpd}
     \end{subfigure}
     \begin{subfigure}[b]{0.48\textwidth}
         \centering
         \includegraphics[width=\textwidth]{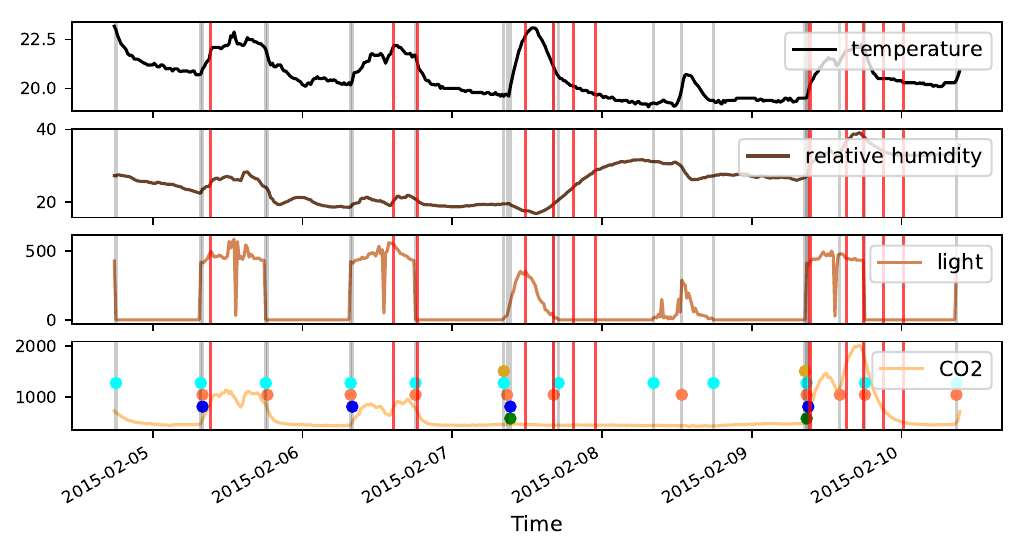}
         \vspace*{-6mm}
         \caption{ALACPDw/oAR}
         \label{fig:occupancy_alacpd_lstm}
     \end{subfigure}
     \begin{subfigure}[b]{0.48\textwidth}
         \centering
         \includegraphics[width=\textwidth]{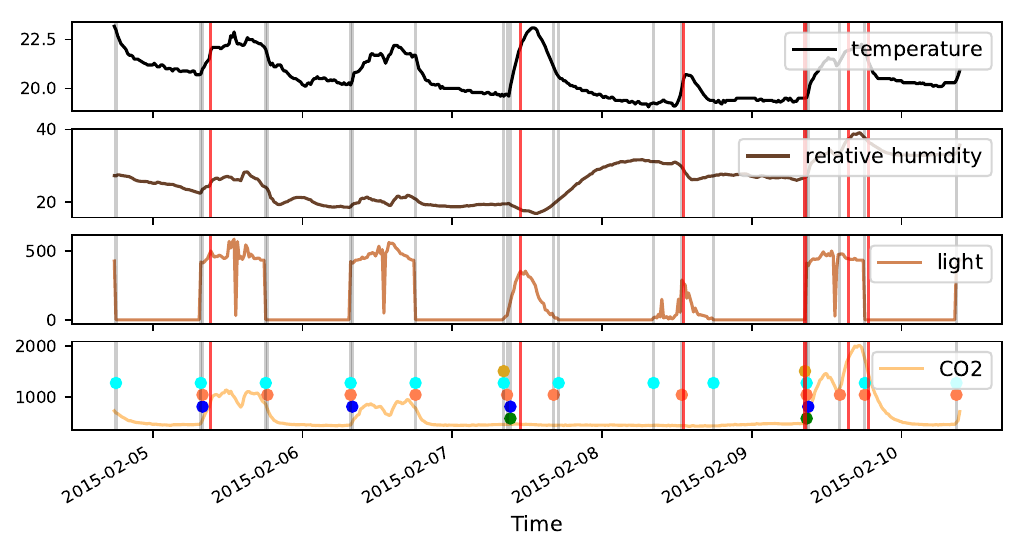}
         \vspace*{-6mm}
         \caption{ALACPDw/oAE}
         \label{fig:occupancy_alacpd_ar}
     \end{subfigure}
     
     \caption{CPD results on the Occupancy dataset. The gray and red vertical lines depict all the change points determined by the annotators and the algorithm, respectively. The colored points on the lines refer to the annotators detected the corresponding change points; each color corresponds to a single annotator.}
        \label{fig:cpd_occupancy}
\end{figure*}

\begin{figure*}[!ht]
     \centering
     
     \begin{subfigure}[b]{0.48\textwidth}
         \centering
         \includegraphics[width=\textwidth]{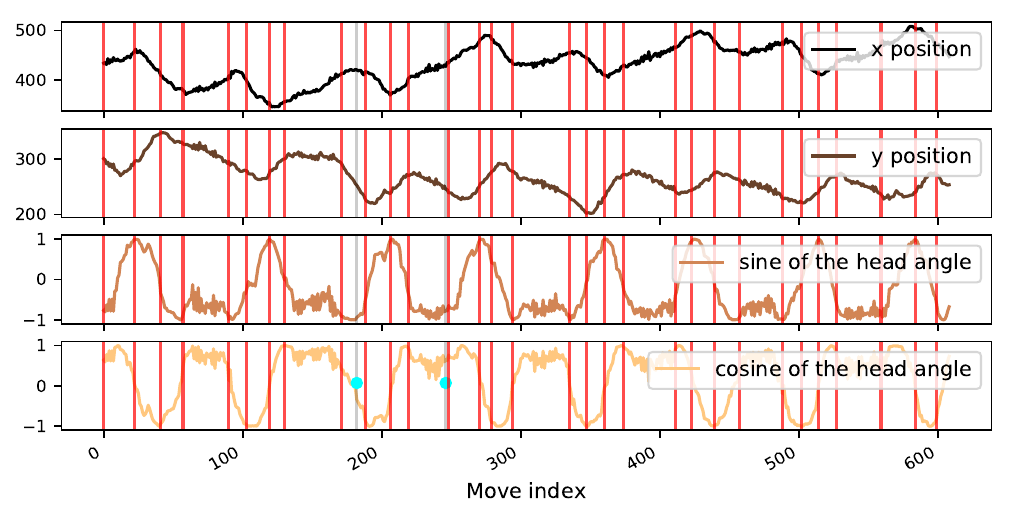}
         \caption{BOCPD}
         \label{fig:bee_waggle_6_bocpd}
     \end{subfigure}
     \begin{subfigure}[b]{0.48\textwidth}
         \centering
         \includegraphics[width=\textwidth]{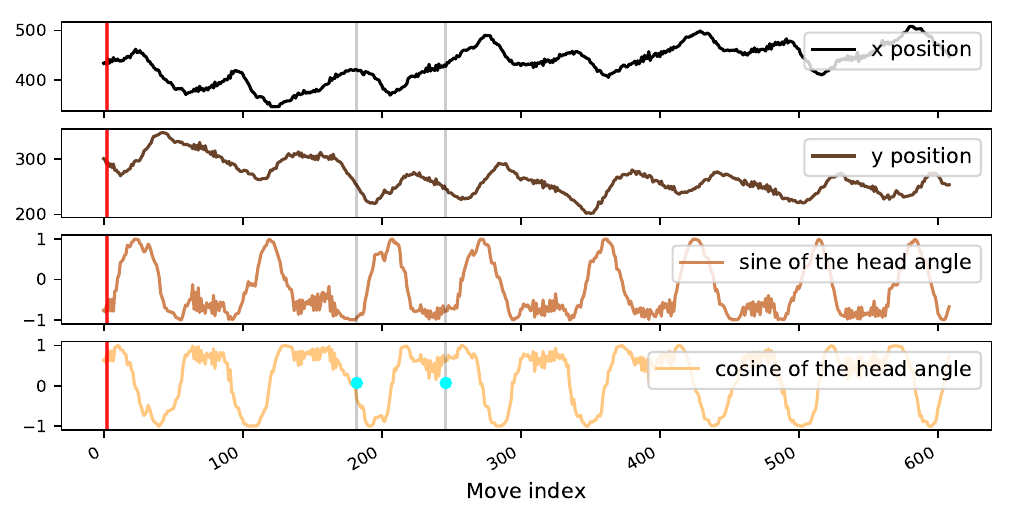}
         \caption{BOCPDMS}
         \label{fig:bee_waggle_6_bocpdms}
     \end{subfigure}
        \
    \begin{subfigure}[b]{0.48\textwidth}
         \centering
         \includegraphics[width=\textwidth]{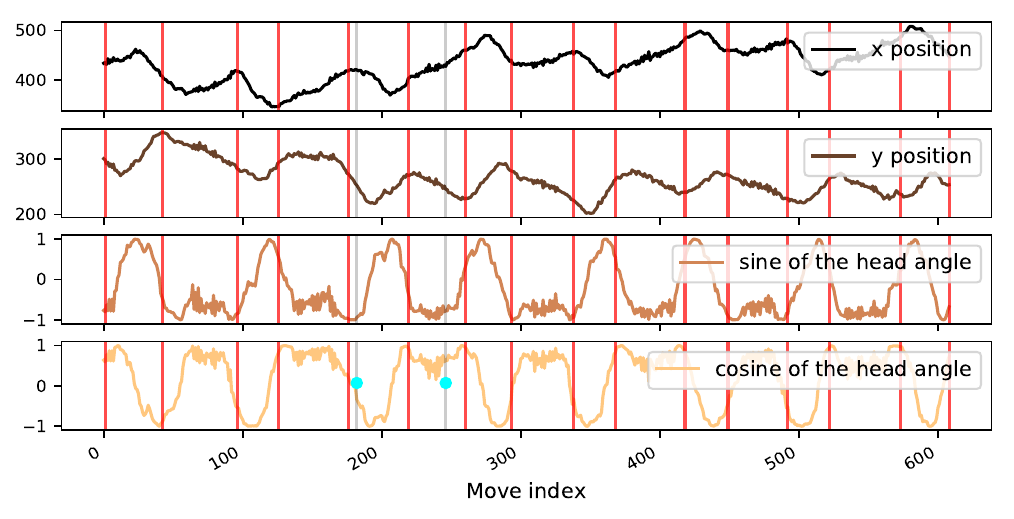}
         \caption{ECP}
         \label{fig:bee_waggle_6_ecp}
     \end{subfigure}
    \begin{subfigure}[b]{0.48\textwidth}
         \centering
         \includegraphics[width=\textwidth]{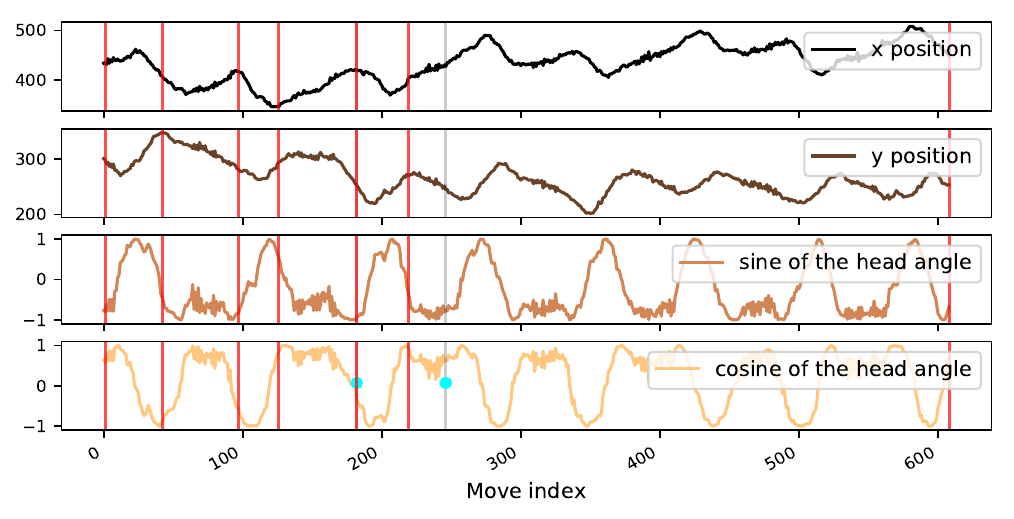}
         \caption{KCPA}
         \label{fig:bee_waggle_6_kcpa}
     \end{subfigure}
     \begin{subfigure}[b]{0.48\textwidth}
         \centering
         \includegraphics[width=\textwidth]{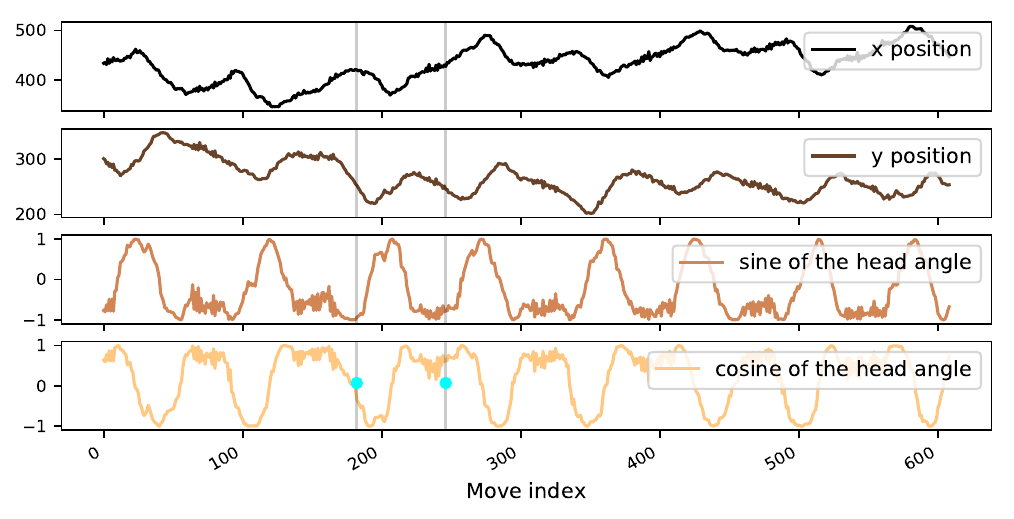}
         \caption{ZERO}
         \label{fig:bee_waggle_6_zero}
     \end{subfigure}
     \begin{subfigure}[b]{0.48\textwidth}
         \centering
         \includegraphics[width=\textwidth]{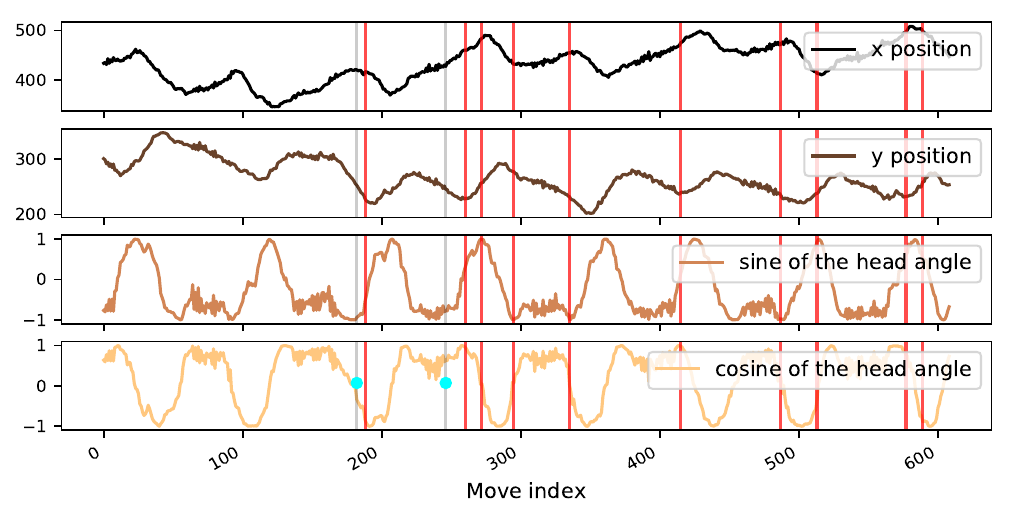}
         \caption{ALACPD}
         \label{fig:bee_waggle_6_alacpd}
     \end{subfigure}
     \begin{subfigure}[b]{0.48\textwidth}
         \centering
         \includegraphics[width=\textwidth]{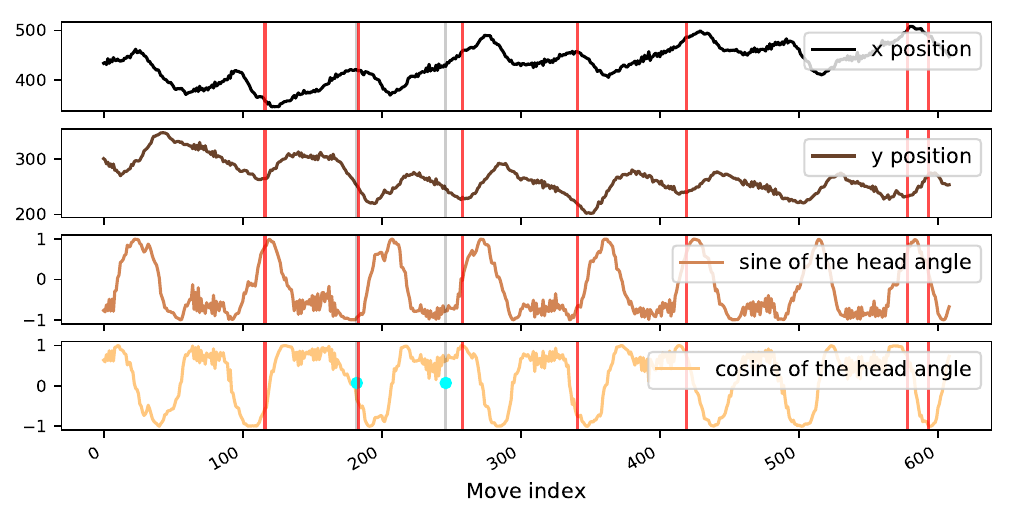}
         \caption{ALACPDw/oAR}
         \label{fig:bee_waggle_6_alacpd_lstm}
     \end{subfigure}
     \begin{subfigure}[b]{0.48\textwidth}
         \centering
         \includegraphics[width=\textwidth]{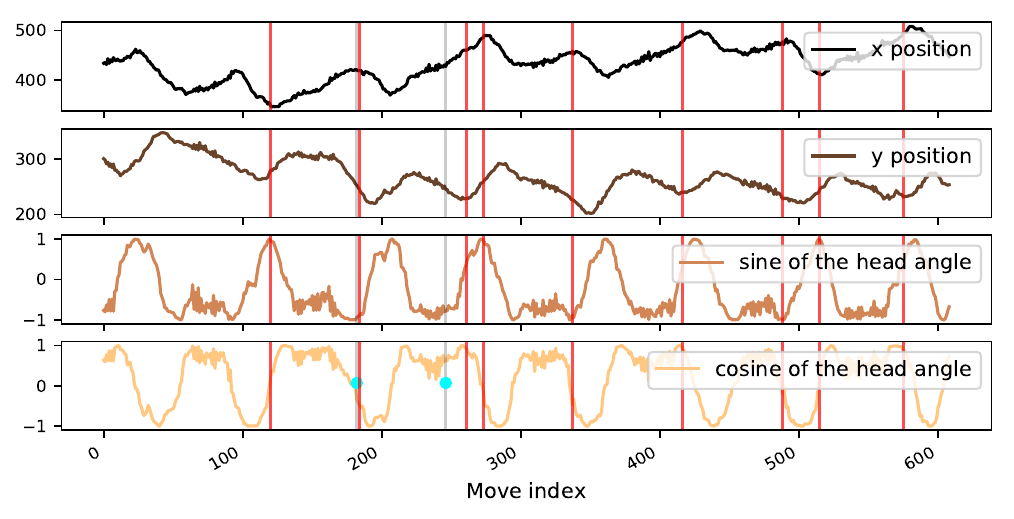}
         \caption{ALACPDw/oAE}
         \label{fig:bee_waggle_6_alacpd_ar}
     \end{subfigure}
     \caption{CPD results on the Bee\_waggle\_6 dataset. The gray and red vertical lines depict all the change points determined by the annotators and the algorithm, respectively. The colored points on the lines refer to the annotators detected the corresponding change points; each color corresponds to a single annotator.}
        \label{fig:cpd_bee_waggle_6}
\end{figure*}

In this appendix, we visualize the detected change points by the considered methods in Figures  \ref{fig:cpd_apple} \ref{fig:cpd_run_log}, \ref{fig:cpd_occupancy}, \ref{fig:cpd_bee_waggle_6}.

\end{appendices}
\end{document}